\documentclass{article}

\usepackage{iclr2027_conference}
\let\cite\citep
\usepackage[T1]{fontenc}
\usepackage{times}
\usepackage{pifont}
\newcommand{\goalcheck}{\ding{51}}
\newcommand{\goalcross}{\ding{55}}

\usepackage{amsmath,amsfonts,bm}

\def\Figref#1{Fig.~\ref{#1}}
\def\cref#1{Sec.~\ref{#1}}
\def\Tabref#1{Tab.~\ref{#1}}

\def\eqref#1{equation~\ref{#1}}
\def\Eqref#1{Eq.~(\ref{#1})}

\def\1{\bm{1}}

\DeclareMathAlphabet{\mathsfit}{\encodingdefault}{\sfdefault}{m}{sl}
\SetMathAlphabet{\mathsfit}{bold}{\encodingdefault}{\sfdefault}{bx}{n}

\DeclareFontShape{OML}{cmm}{b}{it}{<-> cmmib10}{}
\DeclareMathSizes{8}{10}{10}{10}

\usepackage{amsmath}
\usepackage{amssymb}
\usepackage{booktabs}
\usepackage{graphicx}
\usepackage{tikz}
\usetikzlibrary{arrows.meta,positioning,fit,calc,backgrounds}
\usepackage{float}
\floatstyle{ruled}
\newfloat{algorithm}{tbp}{loa}
\floatname{algorithm}{Algorithm}
\floatstyle{plain}
\usepackage{placeins}
\usepackage{microtype}
\usepackage{xcolor}
\definecolor{taskseg}{RGB}{93,169,157}
\definecolor{taskdepth}{RGB}{225,157,62}
\definecolor{taskvos}{RGB}{157,124,184}
\definecolor{taskplace}{RGB}{96,146,195}
\definecolor{taskoverall}{RGB}{194,125,142}
\definecolor{linkblue}{RGB}{0,0,128}
\newcommand{\segcell}[2]{{\setlength{\fboxsep}{0pt}\colorbox{taskseg!#1!white}{\strut\makebox[3.2em]{#2}}}}
\newcommand{\depthcell}[2]{{\setlength{\fboxsep}{0pt}\colorbox{taskdepth!#1!white}{\strut\makebox[3.2em]{#2}}}}
\newcommand{\voscell}[2]{{\setlength{\fboxsep}{0pt}\colorbox{taskvos!#1!white}{\strut\makebox[3.2em]{#2}}}}
\newcommand{\tworoomcell}[2]{{\setlength{\fboxsep}{0pt}\colorbox{taskseg!#1!white}{\strut\makebox[3.2em]{#2}}}}
\newcommand{\mazecell}[2]{{\setlength{\fboxsep}{0pt}\colorbox{taskdepth!#1!white}{\strut\makebox[3.2em]{#2}}}}
\newcommand{\pushcell}[2]{{\setlength{\fboxsep}{0pt}\colorbox{taskvos!#1!white}{\strut\makebox[3.2em]{#2}}}}
\newcommand{\heatcell}[4]{{\setlength{\fboxsep}{0pt}\colorbox{#1!#2!white}{\strut\makebox[#3]{#4}}}}
\usepackage{hyperref}
\hypersetup{
  colorlinks=true,
  linkcolor=linkblue,
  citecolor=linkblue,
  urlcolor=linkblue
}
\usepackage{url}

\title{Slot-JEPA: Object-Structured Prediction Interfaces for \mbox{Latent World Models}}

\author{\normalfont
\parbox{0.95\textwidth}{\centering
Peisong Wen$^{1}$ \quad
Qianqian Xu$^{2,3,*}$ \quad
Minye Lei$^{1,2}$\\[0.35ex]
Siran Dai$^{4,5}$ \quad
Yang Liu$^{1}$ \quad
Qingming Huang$^{1,*}$\\[0.7ex]
{\footnotesize $^{1}$School of Computer Science and Technology, University of Chinese Academy of Sciences}\\
{\footnotesize $^{2}$State Key Laboratory of AI Safety, Institute of Computing Technology, Chinese Academy of Sciences}\\
{\footnotesize $^{3}$Beijing Academy of Artificial Intelligence}\\
{\footnotesize $^{4}$Institute of Information Engineering, Chinese Academy of Sciences}\\
{\footnotesize $^{5}$School of Cyber Security, University of Chinese Academy of Sciences}\\[0.6ex]
{\footnotesize
\begin{tabular}{@{}c@{\qquad}c@{}}
\href{mailto:wenpeisong@ucas.ac.cn,qmhuang@ucas.ac.cn}{\texttt{\{wenpeisong,qmhuang\}@ucas.ac.cn}} &
\href{mailto:xuqianqian@ict.ac.cn,leiminye25z@ict.ac.cn}{\texttt{\{xuqianqian,leiminye25z\}@ict.ac.cn}} \\
\href{mailto:daisiran@iie.ac.cn}{\texttt{daisiran@iie.ac.cn}} &
\href{mailto:liuyang232@mails.ucas.ac.cn}{\texttt{liuyang232@mails.ucas.ac.cn}}
\end{tabular}}\\[0.4ex]
{\small $^*$Corresponding authors.}
}}

\iclrfinalcopy

\begin{document}

\maketitle

\setlength{\abovedisplayskip}{4pt}
\setlength{\belowdisplayskip}{4pt}
\setlength{\abovedisplayshortskip}{2pt}
\setlength{\belowdisplayshortskip}{3pt}

\begin{abstract}
As a prototypical latent world model, Joint-Embedding Predictive Architectures (JEPAs) provide strong semantic representations. However, dense patch representations expose a redundant interface to downstream predictors, which must implicitly infer spatial organization and temporal correspondence, presenting three challenges. First, object information is distributed across patches, forcing each downstream predictor to redundantly recover object structure that could otherwise be reused across tasks before modeling interactions. Second, the lack of explicit entity correspondence across time makes historical transitions expensive, restricting the predictor's flexibility in utilizing context. Third, limited by the lack of explicit transport structure, existing JEPA models typically use pointwise $\ell_1/\ell_2$ distances, which are insensitive to spatial displacement when predicted and goal supports are largely disjoint. We propose the \textbf{Slot-based Joint-Embedding Predictive Architecture (Slot-JEPA)}, which decomposes dense representations into stable semantic slots and time-varying spatial masks. Pretrained once and frozen downstream, this decomposition exposes a reusable object-structured interface. Building on it, our \textbf{Transition Memory Predictor (TMP)} stores visual transitions as compact masks and retrieves useful context using statistics of corresponding slots. Its two branches separate slot-specific dynamics from shared cross-slot dynamics. We further use Unbalanced Optimal Transport (UOT) to supervise mask prediction and evaluate candidate plans by transporting mass between aligned predicted and goal masks, providing a signal better aligned with spatial progress. Across seven visual-understanding benchmarks, the slot-reconstructed features remain transferable. Slot-JEPA also improves goal-conditioned planning across six planar environments and four robotic arm planning tasks on the REALM-JEPA benchmark. Code is available at \url{https://github.com/KID-7391/Slot-JEPA.git}.
\end{abstract}

\section{Introduction}
\label{sec:introduction}

Building agents that can anticipate the consequences of their actions is a long-standing goal of artificial intelligence. World models pursue this goal by predicting how an environment evolves, allowing agents to evaluate possible behaviors before acting \cite{ha2018worldmodels}. Early visual world models operate in pixel space \cite{finn2016physical,finn2017foresight}, but reconstructing every pixel requires modeling details that may be unpredictable. Latent world models instead predict action-conditioned changes in learned representations \cite{hafner2019planet,hafner2020dreamer} without rendering pixels.

Joint-Embedding Predictive Architectures (JEPAs) provide a promising foundation for this approach. Typically, JEPAs learn to predict masked image/video representations from visible context \cite{assran2023ijepa,bardes2024vjepa}. DINO-WM predicts representations changing from offline interactions \cite{zhou2025dinowm}, while V-JEPA 2-AC post-trains an action-conditioned predictor for planning \cite{assran2025vjepa2}. Recent work further improves representation density and temporal consistency \cite{murlabadia2026vjepa21}, or adapts the model online \cite{wang2026adajepa}. Parallel studies diagnose structural degradation and over-dispersed patch semantics in dense self-supervised representations, and develop structure-aware selection or semantic concentration to improve dense transfer \cite{dai2025structuraldegradation,wen2026semantic}. For video representations, T-CoRe reduces temporal sampling uncertainty using two-sided temporal correspondence, while Co-Settle studies the trade-off between intra-video consistency and inter-video separability during image-to-video transfer \cite{liu2025futurepast,liu2026staticdynamic}.

Despite this progress, we argue that dense patch representations might not be ideal for downstream dynamics learning and planning for three reasons: \textbf{1)} An object's information is distributed across patches, leaving its structure implicit. Each task-specific predictor must recover this structure, model interactions, and decode the result back into patches. The object organization encoded during pretraining is not exposed as a reusable interface. \textbf{2)} A predictor needs relevant and diverse context to balance local and global views of dynamics. Yet patch-based history is highly redundant, making transitions expensive to store and their relevance costly to compute. \textbf{3)} Patch representations form dense spatial grids without explicit entity correspondence or transport structure. Existing methods therefore compare predicted and goal grids with pointwise $\ell_1$ or $\ell_2$ distance \cite{zhou2025dinowm,assran2025vjepa2}. When their supports are largely disjoint, these losses can become insensitive to spatial displacement and rank progress poorly.

We therefore propose \textit{\textbf{Slot}-based \textbf{J}oint-\textbf{E}mbedding \textbf{P}redictive \textbf{A}rchitecture (\textbf{Slot-JEPA})}, which turns dense patches into a reusable object-structured interface. A task-agnostic decomposition module addresses problem \textbf{1)} by factorizing the dense feature field into stable semantic slots \cite{locatello2020slotattention,seitzer2023dinosaur,nam2026causaljepa} and time-varying spatial masks. Pretrained once and frozen for downstream tasks, it preserves the original feature field through reconstruction while exposing its object organization.

For problem \textbf{2)}, the temporal stability of slots allows visual history to be stored and reused as compact masks rather than high-dimensional patch representations. Inspired by memory-based Video Object Segmentation \cite{oh2019stm,cheng2024cutie}, our proposed \textit{\textbf{T}ransition \textbf{M}emory \textbf{P}redictor (\textbf{TMP})} uses statistics of masks associated with the same slots to assess relevance to the current query and retrieve relevant and diverse, potentially non-contiguous transitions. We further design two branches that separate slot-specific dynamics from inter-slot interaction dynamics, reducing the burden on dynamics learning.

Finally, problem \textbf{3)} motivates our Unbalanced Optimal Transport (UOT) loss for planning. Using the temporal correspondence of slots, it replaces pointwise comparison between predicted and goal patch representations with transport between their aligned masks. Moving mask mass across space provides a smooth planning signal that remains informative over long displacements.

Slot decomposition turns a perception-oriented patch field into a reusable interface for prediction, memory maintenance, and goal evaluation. Across seven visual-understanding benchmarks, slot-reconstructed features retain most of the semantic information. Across six planar tasks and four Robotic Arm Planning tasks from the REALM-JEPA dataset, Slot-JEPA improves goal-conditioned planning.
The contributions are:
\begin{itemize}
    \item We introduce Slot-JEPA, which compresses patch representations into shared semantic slots and time-varying masks, separating reusable visual organization from task-specific dynamics and easing predictor learning.
    \item We develop TMP to retrieve relevant, non-contiguous transitions from memory instead of relying only on a continuous recent-history window.
    \item We predict action effects in mask space and use UOT for goal evaluation, providing a signal better aligned with spatial progress when $\ell_1/\ell_2$ losses become insensitive to displacement.
\end{itemize}

\section{Related Work}
\label{sec:related-work}

Slot-JEPA connects four lines of work that are often studied separately: joint-embedding representation prediction, object-centric decomposition and dynamics, memory-based video segmentation, and optimal transport. Together, they motivate its slot-and-mask state, transition memory, and spatial planning objective.

\subsection{Joint-Embedding Predictive World Models}
\label{sec:related-work-jepa}

Joint-embedding methods predict target representations rather than reconstructing raw observations, allowing the learned state to retain predictable semantic structure while discarding some pixel-level detail. I-JEPA introduces this principle for masked image regions, and V-JEPA extends it to spatiotemporal video representations \cite{assran2023ijepa,bardes2024vjepa}. Dense self-supervised features from DINO and DINOv2 provide another strong latent target for spatial prediction \cite{caron2021dino,oquab2024dinov2}. Recent work identifies structural degradation over the course of self-supervised training and over-dispersion of semantically related patches, then addresses these issues through structure-aware model selection or explicit semantic concentration \cite{dai2025structuraldegradation,wen2026semantic}. In video self-supervision, T-CoRe uses frames on both sides of the prediction target to reduce temporal sampling uncertainty, while Co-Settle studies image-to-video transfer as a balance between within-video temporal consistency and cross-video semantic separability \cite{liu2025futurepast,liu2026staticdynamic}. These methods primarily concern how a useful visual representation is learned.

A second line of work turns representation prediction into an action-conditioned world model. DINO-WM predicts future DINOv2 patch features from offline interaction trajectories and plans by comparing imagined features with goal features \cite{zhou2025dinowm}. V-JEPA 2-AC post-trains a frame-causal, action-conditioned predictor for robot planning, and V-JEPA 2.1 improves the spatial density and temporal consistency of the underlying video features \cite{assran2025vjepa2,murlabadia2026vjepa21}. Subsequent studies isolate the effects of representation, predictive objective, and planner design \cite{terver2026jepawm}; AdaJEPA adapts the encoder and predictor online from newly observed transitions \cite{wang2026adajepa}; and LeWorldModel investigates stable end-to-end training of JEPA world models from pixels \cite{maes2026leworldmodel}.

Slot-JEPA is complementary to these advances: rather than replacing joint-embedding pretraining, it introduces a structured slot-and-mask interface between a pretrained dense encoder and action dynamics, together with a spatial objective for prediction and planning.

\subsection{Object-Centric Representation and Dynamics}
\label{sec:related-work-object-centric}

Slot Attention maps a set of visual features to a small collection of exchangeable components through competition for input tokens \cite{locatello2020slotattention}. Subsequent methods extend this idea from individual images to video. SIMONe separates temporally abstracted object latents from per-pixel mixture assignments \cite{kabra2021simone}; SAVi and SAVi++ use temporally propagated slots to support object tracking in synthetic and real videos \cite{kipf2022savi,elsayed2022savipp}; and SlotFormer learns dynamics over slot sequences for video prediction and planning \cite{wu2023slotformer}. DINOSAUR shows that reconstructing pretrained self-supervised features, rather than RGB pixels, can induce object-centric components in real images \cite{seitzer2023dinosaur}. This last observation is especially relevant to our goal of reorganizing a pretrained feature field without discarding its semantic information.

Object-centric prediction commonly updates the full latent content of every slot at each time step. Causal-JEPA, the closest joint-embedding counterpart, applies masked and future prediction to object-centric latents \cite{nam2026causaljepa}; Better Slots, Better Worlds studies how the quality of these latents affects prediction, planning, and robustness \cite{nazirjonov2026betterslots}. In contrast, Slot-JEPA holds a reference-derived semantic slot bank fixed within a prediction context and predicts its spatial masks. As in unsupervised slot learning generally, these slots are object-centric components rather than guaranteed physical-instance segmentations.

\subsection{Memory-Based VOS and Transition Memory}
\label{sec:related-work-memory}

Semi-supervised Video Object Segmentation (VOS) must preserve the identity of a target while its appearance and location change. Early propagation methods predict the next mask from an initial annotation and recent frames or masks \cite{perazzi2017masktrack,oh2018rgmp}. Space-Time Memory Networks instead store past image--mask pairs and let the current query read from them \cite{oh2019stm}. XMem separates working, long-term, and sensory memories for long videos, while adaptive memory methods retain frames according to their utility rather than keeping every observation \cite{cheng2022xmem,pourganjalikhan2022adaptive}. Cutie introduces object-level memory reading, and SAM~2 develops a streaming memory architecture for promptable segmentation \cite{cheng2024cutie,ravi2024sam2}. Collectively, these methods show that a query can benefit from selected non-contiguous history instead of only the immediately preceding frame.

TMP transfers this memory principle from perceptual propagation to controlled prediction. Unlike VOS memories that store evidence for locating a target, TMP stores observed source--action--target transitions and retrieves potentially non-contiguous examples to predict a new action outcome.

\subsection{Unbalanced Optimal Transport for Visual Correspondence and Control}
\label{sec:related-work-uot}

Pointwise distances compare features or masks only at matching grid locations. They can therefore provide little information about \emph{how far} mass should move when predicted and target supports do not overlap. Optimal transport instead compares nonnegative spatial measures under a ground cost, explicitly charging for displacement between source and target locations. Entropic regularization makes the problem amenable to efficient Sinkhorn scaling \cite{cuturi2013sinkhorn}. Unbalanced optimal transport (UOT) further relaxes exact marginal matching by assigning a finite penalty to the creation or removal of mass \cite{chizat2018uot,chizat2018scaling}. This relaxation is appropriate for visual masks, whose total mass can change with occlusion, viewpoint, scale, or imperfect decomposition.

Slot-JEPA applies UOT between masks with the same slot index, using their shared correspondence to preserve semantic grouping. UOT complements pointwise and reconstruction losses during training and supplies a displacement-aware goal cost during planning; its ground cost can additionally incorporate obstacle geometry in planar environments.

\section{Methodology}
\label{sec:method}

\subsection{Overview}
\label{sec:overview}

At each time step, the multimodal observation $o_t=(v_t,r_t)$ contains a visual observation $v_t$ and, when available, a low-dimensional state observation $r_t$, such as the robot end-effector pose and gripper state; $a_t$ denotes the action. Given previously observed sequences, a world model predicts how the world will evolve under candidate future action sequences. At test time, given a current observation $o_0$ and a visual goal $v_g$, model-predictive control (MPC) \cite{chua2018pets} selects the action sequence whose predicted visual observations are closest to $v_g$, as measured by a goal distance $d(\hat v,v_g)$.

\begin{figure}[!tp]
\centering
\includegraphics[width=1.0\linewidth]{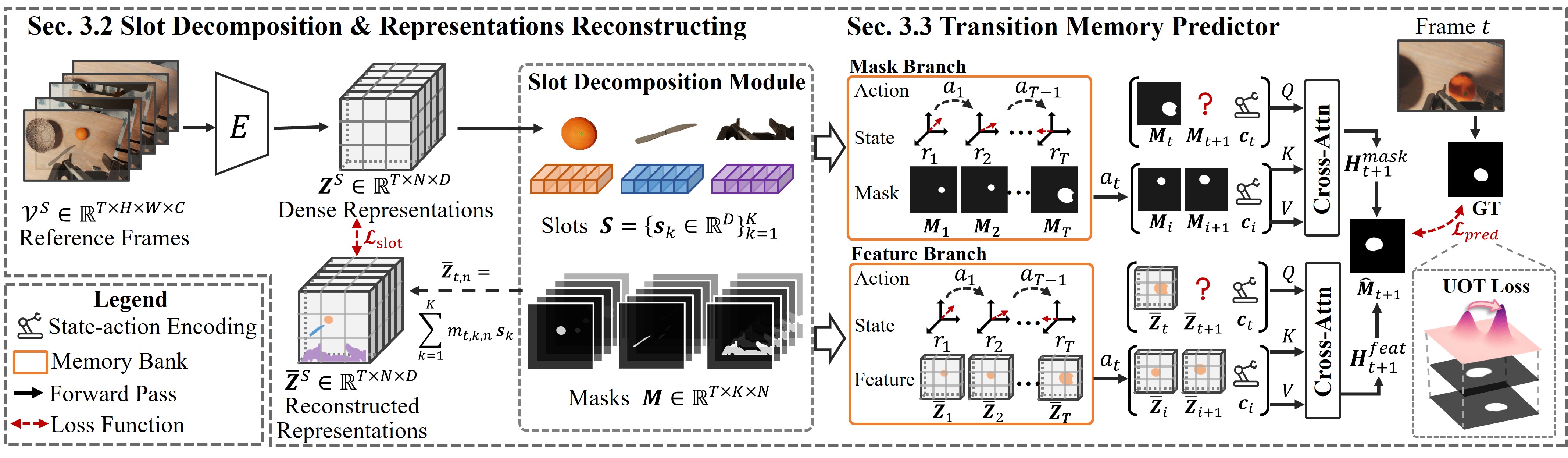}
\vspace{-6mm}
\caption{Overview of Slot-JEPA. The slot decomposition module factorizes patch representations into stable slots $\bm{S}$ and time-varying masks $\bm{M}$. TMP predicts $\hat{\bm{M}}_{t+1}$ with sampled transition memory. The prediction is supervised against the next-frame mask by $\mathcal{L}_{\mathrm{pred}}$, including UOT loss.}
\label{fig:overview}
\end{figure}

An overview of Slot-JEPA is shown in \Figref{fig:overview}. Slot-JEPA inserts a reusable object-structured interface between a general visual encoder and task-specific predictors. We therefore initialize the visual encoder with V-JEPA 2.1 \cite{murlabadia2026vjepa21}, and jointly pretrain the slot decomposition module and Transition Memory Predictor (TMP). The module converts patch representations into stable slots and spatial masks, which serve as the basic TMP inputs and outputs. After pretraining, we freeze the visual encoder and slot decomposition module, and fine-tune the transferred TMP on each downstream control task. A UOT-based loss is applied during training and planning.

\subsection{Pretrained Slot Decomposition Module}
\label{sec:slot-decomposition}

Let $\mathcal{V}^{S}$ denote the reference frames for slot extraction, and let $\bm{Z}^{S}=E(\mathcal{V}^{S})$ be their dense patch representations. The slot decomposition module extracts a shared set of $K$ slots $\bm{S}=\{\bm{s}_k\in\mathbb{R}^{D}\}_{k=1}^K$ from $\bm{Z}^{S}$. Holding $\bm{S}$ fixed, it maps any encoded sequence $\bm{Z}=E(v_{1:T})\in\mathbb{R}^{T\times N\times D}$ from the same environment to a time-varying spatial assignment $\bm{M}=[m_{t,k,n}]\in[0,1]^{T\times K\times N}$. The slots summarize visual content shared by the reference frames, while $m_{t,k,:}\in[0,1]^N$ indicates where the content of slot $k$ appears. Ideally, they reconstruct each patch representation through
\begin{equation}
\label{eq:slot-reconstruction}
\bm{Z}_{t,n}\approx\bar{\bm{Z}}_{t,n}=\sum_{k=1}^{K}m_{t,k,n}\bm{s}_k.
\end{equation}
Whereas the dense representation $\bm{Z}$ contains $TND$ scalars, the factorized representation $(\bm{S},\bm{M})$ contains $KD+TKN$ scalars. The latter is therefore more compact since $K\ll D$ in practice.
To obtain $\bm{S}$, we use Slot Attention \cite{locatello2020slotattention}, which iteratively aggregates the reference patches into competing slots. Let $\bm{s}^{\mathrm{loc}}_{j,r}$ denote local slot $j$ at iteration $r$, the local update is
\begin{equation}
\begin{aligned}
\alpha_{n,j,r}&=\operatorname{softmax}_{j}\!\left(
k(\bm{z}_n)^\top q(\bm{s}^{\mathrm{loc}}_{j,r}) / \big(\tau\sqrt{D}\big)\right),\quad
\bm{u}_{j,r}=\sum_n\frac{\alpha_{n,j,r}}{\sum_{n'}\alpha_{n',j,r}}v(\bm{z}_n),\\
\tilde{\bm{s}}^{\mathrm{loc}}_{j,r+1}
&=\operatorname{GRU}(\bm{u}_{j,r},\bm{s}^{\mathrm{loc}}_{j,r}),\quad
\bm{s}^{\mathrm{loc}}_{j,r+1}=\tilde{\bm{s}}^{\mathrm{loc}}_{j,r+1}
+\operatorname{MLP}\!\left(\operatorname{LN}(\tilde{\bm{s}}^{\mathrm{loc}}_{j,r+1})\right).
\end{aligned}
\end{equation}
Here, $\tau>0$ controls assignment softness, and GRU, LN, and MLP denote the gated recurrent unit, layer normalization, and multilayer perceptron, respectively. For initialization, we partition the reference frames into non-overlapping windows, extract local slots from each window, and use farthest-point sampling to select diverse initial slots. 
Importantly, Slot-JEPA does not assume that slots correspond to predefined semantic categories. Instead, the shared reference-derived slots provide a fixed latent basis within each planning context, and masks encode temporal variation relative to this basis.

On top of the extracted slots $\bm{S}$, we compute $\bm{M}$ in the pretrained latent space \cite{kabra2021simone}. Patch features cross-attend to the slots, and an MLP $g$ produces slot-assignment logits:
\begin{equation}
\bm{H}=\operatorname{CA}(\bm{Z};\bm{S}),\qquad
A_{t,k,n}=g\!\left([\bm{H}_{t,n}\Vert \bm{s}_k]\right),\qquad
m_{t,k,n}=\operatorname{softmax}_{k}(A_{t,k,n}).
\end{equation}
Here, $\operatorname{CA}$ denotes the cross-attention stack. Competition across slots encourages object-aligned grouping but does not require one slot per physical instance: a slot may cover multiple entities, a part, or background. Increasing $K$ permits a finer decomposition but increases the prediction cost.

Having obtained $\bm{S}$ and $\bm{M}$ for frame $t$, we reconstruct the patch representation $\bar{\bm{Z}}_t$ with \Eqref{eq:slot-reconstruction}. Then we train the decomposition module by reconstructing the original dense representation $\bm{Z}_t$, while encouraging different slots to encode distinct content:
\begin{equation}
\mathcal{L}_{\mathrm{slot}}=\frac{1}{T}\sum_{t=1}^T\lVert\bar{\bm{Z}}_t-\bm{Z}_t\rVert_1+\frac{\lambda_{\mathrm{orth}}}{K(K-1)}\sum_{k\neq j}\Big(\frac{\bm{s}_k^\top\bm{s}_j}{\lVert\bm{s}_k\rVert_2\lVert\bm{s}_j\rVert_2}\Big)^2.
\end{equation}

\subsection{Transition Memory Predictor}
\label{sec:tmp}

The temporally stable slots make compact, query-dependent reuse of the observed trajectory possible: visual states can be stored as masks and compared through statistics of corresponding slots. During training, the observed transitions in each sampled video clip provide the context directly. At inference, TMP samples relevant and diverse context transitions $\mathcal{C}_t$ based on the query state and mask geometry. It then uses the same symmetric context--query structure to predict the next mask in both settings.

\paragraph{Transition memory bank.}
Each memory bank contains transitions from one environment that are decoded against the same slots $\bm{S}$. Visual states can therefore be stored solely as masks $\bm{M}_i$: 
\begin{equation}
\mathcal{B}_t=\{(\bm{M}_i,r_i,a_i,\bm{M}_{i+1},r_{i+1})\}_{i\in\mathcal{I}^{\mathrm{mem}}_t},
\end{equation}
where $r_i$ is the optional low-dimensional state, $a_i$ is the action, and $\mathcal{I}^{\mathrm{mem}}_t$ indexes the transitions available before the query. 
Let $\bm{\rho}_i\in\mathbb{R}^{K}$, $\bm{\mu}_i\in\mathbb{R}^{2K}$ collect the areas of the $K$ masks and 2-D centroids in $\bm{M}_i$ respectively, and $\bm{u}_i=[ r_i\Vert\sqrt{\lambda_v}\bm{\rho}_i\Vert\sqrt{\lambda_v}\bm{\mu}_i]$ summarizes each state. At inference, we sample transitions $\mathcal{C}_t\subset\mathcal{B}_t$ without replacement according to
\begin{equation}
d_{\mathrm{mem}}(t,i)=\lVert\bm{u}_t-\bm{u}_i\rVert_2^2,\qquad P_t(i)=(1-\gamma)\operatorname{softmax}_i\!\left(-d_{\mathrm{mem}}(t,i)\right)+\gamma / |\mathcal{B}_t|.
\label{eq:context-sampling}
\end{equation}
The mixing weight $\gamma\in[0,1]$ balances distance-weighted sampling for relevance with uniform sampling for diversity, which is essential for in-context examples \cite{huang2026backfires}. 

\paragraph{Predictor with symmetric structure.}
TMP contains a slot-wise mask branch for spatial changes and a shared feature branch for interactions among slots. Both process observed transitions as context blocks and the transition to be predicted as a query block.
For each context transition $i$, we encode its state and action as $\bm{c}_i=E_c([r_i\Vert a_i \Vert r_{i+1}])$. In the mask branch, slot $k$ forms a spatial field $\bm{X}_{i,k,n}=m_{i,k,n}\bm{s}_k$. The current and next fields are paired to form a context block
\begin{equation}
\bm{B}_{i,k}=\big[P_c(\bm{c}_i),P_s(\bm{s}_k),P_m(\bm{X}_{i,k}\Vert\bm{X}_{i+1,k})\big]\in\mathbb{R}^{(N'+2)\times d'}.
\end{equation}
Here $P_c$, $P_s$, and $P_m$ project inputs to width $d'$. For the query, unknown future inputs are replaced by learnable tokens. Specifically, $\bm{q}^{r}$ replaces $r_{t+1}$ in the state--action encoding: $\bm{c}_t=E_c([r_t\Vert a_t\Vert\bm{q}^{r}])$, and $\bm{q}_k$ replaces the future spatial field. Then we have
\begin{equation}
\bm{B}_{t,k}=\big[P_c(\bm{c}_t),P_s(\bm{s}_k),P_m(\bm{X}_{t,k}\Vert\bm{q}_k)\big],\quad
\bm{B}_k=\big[\bm{B}_{i,k}\big]_{i\in\mathcal{I}_t},\quad
\bm{B}_k^+=[\bm{B}_k;\bm{B}_{t,k}].
\end{equation}
where $\mathcal{I}_t$ indexes the selected transitions. Then we apply multiple attention blocks:
\begin{equation}
\bm{H}_{t,k}^{\mathrm{mask}}=\operatorname{Attn}\!\left(\bm{B}_{t,k}W_Q,\bm{B}_k^+W_K,\bm{B}_k^+W_V\right), \quad
\operatorname{Attn}(Q,K,V)=\operatorname{softmax}(QK^\top/\sqrt d)V.
\end{equation}
We select only the query positions $\mathcal{J}$ occupied by learnable tokens and concatenate $(\bm{H}_{t,k}^{\mathrm{mask}})_{\mathcal{J}}$ into $\bm{H}^{\mathrm{mask}}_{t+1}$.
The feature branch follows the same process, but takes $[\bar{\bm{Z}}_i\Vert\bm{M}_i^\top]$ as its spatial input, yielding interaction features $\bm{H}^{\mathrm{feat}}_{t+1}$.
Finally, we broadcast $\bm{H}^{\mathrm{feat}}_{t+1}$ across slots and combine both branches with an MLP $f_{\mathrm{out}}$ and slot-wise softmax to predict the future assignment map:
\begin{equation}
\hat{\bm{M}}_{t+1}=\operatorname{softmax}_{\mathrm{slot}}\!\left(f_{\mathrm{out}}\!\left([\bm{H}^{\mathrm{mask}}_{t+1}\Vert\bm{H}^{\mathrm{feat}}_{t+1}]\right)\right).
\end{equation}

\paragraph{Action-free pretraining.}
To pretrain TMP on natural videos without action annotations, an inverse transition encoder $B_\psi$ compresses each observed change into a low-dimensional pseudo-action:
\begin{equation}
\tilde a_t=B_\psi(\bm{S},\bm{M}_t,\bm{M}_{t+1})\in\mathbb{R}^{d_b},\qquad d_b\ll KN.
\end{equation}
For the query transition, $\bm{M}_{t+1}$ reaches TMP only through the bottleneck $\tilde a_t$, limiting direct target leakage while preserving the transition information needed for prediction.
After representation pretraining, we discard the inverse transition and state encoders, and replace them with task-specific encoders for $r_t$ and $a_t$. The remaining TMP weights are fine-tuned on downstream interaction data.

\subsection{Transport-Based Objective for Mask Prediction}
\label{sec:uot}

We next specify how the mask predictions are trained and used for goal evaluation. During training, the fixed $\bm{S}$ decodes the held-out future representation $\bm{Z}_{t+1}$ into the target mask $\bm{M}_{t+1}$. Given action $a_t$ and transition context $\mathcal{C}_t$, TMP produces $\hat{\bm{M}}_{t+1}$. The target and predicted masks reconstruct $\bar{\bm{Z}}_{t+1,n}=\sum_k m_{t+1,k,n}\bm{s}_k$ and $\hat{\bm{Z}}_{t+1,n}=\sum_k\hat m_{t+1,k,n}\bm{s}_k$, respectively.

Conventional latent predictors directly predict $\hat{\bm{Z}}_{t+1}$ and minimize its $\ell_1$ distance to $\bm{Z}_{t+1}$ \cite{bardes2024vjepa,assran2025vjepa2}. With our factorization, the same objective naturally transfers to $\lVert\hat{\bm{M}}_{t+1}-\bm{M}_{t+1}\rVert_1$. However, in either representation or mask space, pointwise $\ell_1$ can become insensitive to further spatial displacement when predicted and target supports are largely disjoint. We therefore use \textit{\textbf{U}nbalanced \textbf{O}ptimal \textbf{T}ransport (\textbf{UOT})} to measure the cost of moving mask mass from a prediction to its target.

Consider masks $p,q\in\mathbb{R}_+^N$, let $\bm{\xi}_i\in[0,1]^2$ be the normalized coordinate of patch $i$, and set $c_{ij}=\lVert\bm{\xi}_i-\bm{\xi}_j\rVert_2^2$ in the ground-cost matrix $\bm{C}$. A transport plan $\bm{\pi}\in\mathbb{R}_+^{N\times N}$ moves $\pi_{ij}$ units of mass from patch $i$ to $j$. Because $p$ and $q$ may have different total mass, KL penalties on the plan's marginals allow mass to be created or removed \cite{chizat2018uot}. We obtain an entropy-regularized plan and evaluate its transport and mass-change costs:
\begin{equation}
\label{eq:uot-objective}
\begin{aligned}
\bm{\pi}^{*}_{\varepsilon,\rho}&=\arg\min_{\bm{\pi}\ge 0}
\langle\bm{\pi},\bm{C}\rangle+\varepsilon\,\Omega(\bm{\pi})
+\rho\,\mathrm{KL}(\bm{\pi}\bm{1}\,\|\,p)+\rho\,\mathrm{KL}(\bm{\pi}^\top\bm{1}\,\|\,q),\\
\mathcal{T}_{\varepsilon,\rho}(p,q)&=\langle\bm{\pi}^{*}_{\varepsilon,\rho},\bm{C}\rangle
+\rho\,\mathrm{KL}(\bm{\pi}^{*}_{\varepsilon,\rho}\bm{1}\,\|\,p)
+\rho\,\mathrm{KL}((\bm{\pi}^{*}_{\varepsilon,\rho})^\top\bm{1}\,\|\,q).
\end{aligned}
\end{equation}
Here, $\Omega$ is the entropic regularizer, $\varepsilon$ controls the smoothness of the computed plan, and $\rho$ controls the cost of changing mass. We solve for the plan with Sinkhorn iterations \cite{chizat2018scaling} (see Appendix \ref{sec:appendix-uot-computation}). Combining UOT, mask $\ell_1$, and the original representation reconstruction gives
\begin{equation}
\mathcal{L}_{\mathrm{pred}}={}\lambda_{\mathrm{uot}}\sum_k \mathcal{T}_{\varepsilon,\rho}(\hat m_{t+1,k},m_{t+1,k})+\lambda_{\mathrm{mask}}\lVert\hat{\bm{M}}_{t+1}-\bm{M}_{t+1}\rVert_1
+\lambda_{\mathrm{feat}}\lVert\hat{\bm{Z}}_{t+1}-\bm{Z}_{t+1}\rVert_1.
\end{equation}
At test time, the first two terms form the planning objective after replacing the target masks with $\bm{M}^g$ extracted from the visual goal.
Both the current observation and goal are decoded against the same $\bm{S}$: 
The goal therefore does not receive an independent slot set, and slot index $k$ remains aligned across current, predicted, and goal masks. 
We roll out candidate action sequences, optimize them with the cross-entropy method (CEM) \cite{chua2018pets}, execute an action chunk, and replan. Appendix~\ref{sec:appendix-uot-computation} details generalized Sinkhorn scaling, change-aware weighting, and static-slot identification; Appendix~\ref{sec:appendix-planar} gives the environment-specific obstacle-aware ground cost and MPC settings.

\section{Experiments}
\label{sec:experiments}

\subsection{Experimental Setup}
\label{sec:experimental-setup}

We conduct experiments around four questions: \textbf{1)} Does the slot decomposition preserve the semantic information (\cref{sec:semantic-preservation})? \textbf{2)} How well does Slot-JEPA perform on visual planning in controlled planar environments (\cref{sec:planar})? \textbf{3)} Does the framework generalize to Robotic Arm Planning (\cref{sec:robot})? \textbf{4)} How does each component contribute to performance (\cref{sec:ablation})?

\paragraph{Representation pretraining.}
Starting from a V-JEPA 2.1 encoder \cite{murlabadia2026vjepa21}, we pretrain the slot decomposition module and TMP on Kinetics-400 videos \cite{kay2017kinetics} using inverse transition encoders. We freeze the pretrained visual encoder and slot decomposition module for downstream tasks. The full pretraining setup is given in Appendix~\ref{sec:appendix-pretraining}.

\subsection{Preservation of Semantic Information}
\label{sec:semantic-preservation}

For each image or video frame, the frozen visual encoder and the slot decomposition module jointly provide a patch representation for downstream visual understanding tasks.
We evaluate linear semantic segmentation on COCOStuff-27 \cite{caesar2018cocostuff}, ADE20K \cite{zhou2017ade20k}, and PASCAL VOC \cite{everingham2010voc} to measure category-level structure; monocular depth estimation on NYUv2 \cite{silberman2012nyuv2} and KITTI \cite{geiger2012kitti}, plus NYUv2-to-SUNRGBD transfer \cite{song2015sunrgbd} without adaptation; and VOS on DAVIS-2017 \cite{ponttuset2017davis} using zero-shot label propagation.
The evaluation settings are detailed in Appendix~\ref{sec:appendix-visual}.

\begin{table}[!tp]
\vspace{-3mm}
\caption{Visual-understanding results on all three probes.}
\label{tab:visual-summary}
\centering\footnotesize
\setlength{\tabcolsep}{3pt}
\begin{NoHyper}
\resizebox{\linewidth}{!}{%
\begin{tabular}{lccccccc}
\toprule
\raisebox{-7pt}[0pt][0pt]{Method} & \multicolumn{3}{c}{Segmentation (mIoU$\uparrow$)} & \multicolumn{3}{c}{Depth ($\delta_1\uparrow$)} & VOS ($\mathcal{J}\&\mathcal{F}\uparrow$) \\
\cmidrule(lr){2-4}\cmidrule(lr){5-7}\cmidrule(l){8-8}
& COCO & ADE & VOC & NYUv2 & KITTI & SUNRGBD & DAVIS \\
\midrule
MoCo v3 \cite{chen2021mocov3} & \segcell{0}{43.4} & \segcell{0}{24.4} & \segcell{0}{61.7} & \depthcell{0}{69.8} & \depthcell{0}{77.2} & \depthcell{0}{50.6} & \voscell{44}{61.7} \\
iBOT \cite{zhou2022ibot} & \segcell{33}{52.2} & \segcell{18}{34.3} & \segcell{36}{73.3} & \depthcell{32}{77.7} & \depthcell{59}{88.3} & \depthcell{43}{69.1} & \voscell{55}{64.3} \\
Mugs \cite{zhou2022mugs} & \segcell{26}{51.4} & \segcell{21}{34.6} & \segcell{26}{72.7} & \depthcell{0}{74.9} & \depthcell{55}{87.7} & \depthcell{26}{66.6} & \voscell{46}{62.1} \\
MAE \cite{he2022mae} & \segcell{0}{37.9} & \segcell{0}{20.6} & \segcell{0}{52.3} & \depthcell{0}{65.3} & \depthcell{31}{84.3} & \depthcell{0}{59.1} & \voscell{0}{48.9} \\
DINOv2\textsuperscript{$\dagger$} \cite{oquab2024dinov2} & \segcell{40}{53.0} & \segcell{30}{35.5} & \segcell{62}{74.9} & \depthcell{34}{78.1} & \depthcell{57}{88.0} & \depthcell{36}{68.0} & \voscell{55}{64.2} \\
V-JEPA 2.1 \cite{murlabadia2026vjepa21} & \segcell{38}{52.8} & \segcell{30}{35.5} & \segcell{17}{72.1} & \depthcell{49}{80.5} & \depthcell{65}{\textbf{89.2}} & \depthcell{49}{70.0} & \voscell{62}{65.9} \\
\midrule
\textbf{Slot-JEPA} & \segcell{65}{\textbf{56.1}} & \segcell{65}{\textbf{38.9}} & \segcell{65}{\textbf{75.1}} & \depthcell{65}{\textbf{83.1}} & \depthcell{55}{87.8} & \depthcell{65}{\textbf{72.3}} & \voscell{65}{\textbf{66.6}} \\
\bottomrule
\end{tabular}}
\end{NoHyper}
\end{table}

Results in \Tabref{tab:visual-summary} support two conclusions: \textbf{1) Slot decomposition preserves the main task-relevant semantics under strong compression.} With $K=16$, the factorized slot representation contains only about $2.5\%$ as many scalars as the dense patch field, yet the slot-reconstructed representation retains the semantic information needed across segmentation, depth, and VOS. \textbf{2) Slot reconstruction makes the patch representation more object-aware.} During representation pretraining, the slot-reconstruction objective fine-tunes the visual encoder and propagates object-level grouping back into its patch features. This supplies explicit object-aware supervision that is absent from the original V-JEPA 2.1 training objective, helping explain the consistent downstream gains.

\subsection{Planar Visual Planning}
\label{sec:planar}

\paragraph{Evaluation protocol.}
We evaluate three in-distribution (ID) tasks and their out-of-distribution (OOD) variants. The ID tasks use the same layouts and object shapes during training and testing, whereas the OOD variants use previously unseen layouts or object shapes. In \textit{TwoRoom} (also called \textit{Wall} in prior work) and \textit{Maze} \cite{zhou2025dinowm}, a point agent passes through a doorway and a ball navigates a U-shaped corridor, respectively. Both use position and velocity as $r_t$, while $a_t$ specifies velocity in TwoRoom and force in Maze. In \textit{PushT} \cite{florence2022ibc,chi2023diffusionpolicy}, a pusher moves a T-shaped object and itself to their target poses; $r_t$ is the pusher state and $a_t$ is a relative target tracked by a PD controller. The OOD variants \textit{TwoRoomR} and \textit{MazeR} randomize obstacle layouts, while \textit{PushObj} changes the object shape. Each task provides an initial image $v_0$ and visual goal $v_g$. For comparison, we report DINO-WM baselines that additionally use the low-dimensional goal state for planning in TwoRoom and PushT. We evaluate 50 cases per environment using MPC \cite{rawlings2026mpc} with CEM \cite{deboer2005cem}.
See Appendix~\ref{sec:appendix-planar} for more details.

\begin{table}[!tp]
\caption{Planar-planning success rates (\%). Goal State indicates whether the planner also receives the low-dimensional goal state.}
\label{tab:planar-results}
\centering
\small
\setlength{\tabcolsep}{2.5pt}
\begin{NoHyper}
\resizebox{0.85\linewidth}{!}{%
\begin{tabular}{@{}lccccccc@{}}
\toprule
\raisebox{-7pt}[0pt][0pt]{Method} & \raisebox{-7pt}[0pt][0pt]{Goal State} & \multicolumn{3}{c}{ID Environments} & \multicolumn{3}{c}{OOD Environments} \\
\cmidrule(lr){3-5}\cmidrule(l){6-8}
& & TwoRoom & Maze & PushT & TwoRoomR & MazeR & PushObj \\
\midrule
\raisebox{-5.5pt}[0pt][0pt]{DINO-CLS \cite{caron2021dino}} & \goalcross & \tworoomcell{0}{6} & \mazecell{25}{88} & \pushcell{0}{20} & \tworoomcell{0}{34} & \mazecell{23}{70} & \pushcell{0}{2} \\
& \goalcheck & \tworoomcell{49}{84} & -- & \pushcell{0}{28} & \tworoomcell{23}{48} & -- & \pushcell{0}{8} \\
\raisebox{-5.5pt}[0pt][0pt]{ResNet \cite{he2016resnet}} & \goalcross & \tworoomcell{0}{6} & \mazecell{58}{98} & \pushcell{0}{34} & \tworoomcell{0}{12} & \mazecell{20}{68} & \pushcell{0}{4} \\
& \goalcheck & \tworoomcell{49}{84} & -- & \pushcell{15}{40} & \tworoomcell{35}{60} & -- & \pushcell{0}{16} \\
\raisebox{-5.5pt}[0pt][0pt]{R3M \cite{nair2023r3m}} & \goalcross & \tworoomcell{0}{38} & \mazecell{65}{\textbf{100}} & \pushcell{0}{16} & \tworoomcell{0}{8} & \mazecell{0}{32} & \pushcell{0}{2} \\
& \goalcheck & \tworoomcell{55}{90} & -- & \pushcell{17}{42} & \tworoomcell{21}{46} & -- & \pushcell{36}{36} \\
\raisebox{-5.5pt}[0pt][0pt]{DINO-WM \cite{zhou2025dinowm}} & \goalcross & \tworoomcell{15}{50} & \mazecell{65}{\textbf{100}} & \pushcell{48}{74} & \tworoomcell{57}{82} & \mazecell{49}{86} & \pushcell{26}{28} \\
& \goalcheck & \tworoomcell{49}{84} & -- & \pushcell{61}{88} & \tworoomcell{57}{82} & -- & \pushcell{41}{40} \\
\raisebox{-5.5pt}[0pt][0pt]{V-JEPA 2.1 \cite{murlabadia2026vjepa21}} & \goalcross & \tworoomcell{15}{50} & \mazecell{45}{94} & \pushcell{32}{58} & \tworoomcell{25}{50} & \mazecell{49}{86} & \pushcell{49}{46} \\
& \goalcheck & \tworoomcell{47}{82} & -- & \pushcell{40}{66} & \tworoomcell{45}{72} & -- & \pushcell{60}{54} \\
LeWM \cite{maes2026leworldmodel} & \goalcross & \tworoomcell{45}{80} & \mazecell{58}{98} & \pushcell{61}{88} & \tworoomcell{27}{52} & \mazecell{52}{88} & \pushcell{28}{30} \\
\midrule
\textbf{Slot-JEPA (Ours)} & \goalcross & \tworoomcell{65}{\textbf{100}} & \mazecell{65}{\textbf{100}} & \pushcell{65}{\textbf{92}} & \tworoomcell{65}{\textbf{90}} & \mazecell{65}{\textbf{96}} & \pushcell{65}{\textbf{58}} \\
\bottomrule
\end{tabular}}
\end{NoHyper}
\end{table}

\begin{figure}[!t]
\centering
\begin{minipage}[t]{0.40\linewidth}
\vspace{-10pt}
\centering
\includegraphics[width=\linewidth]{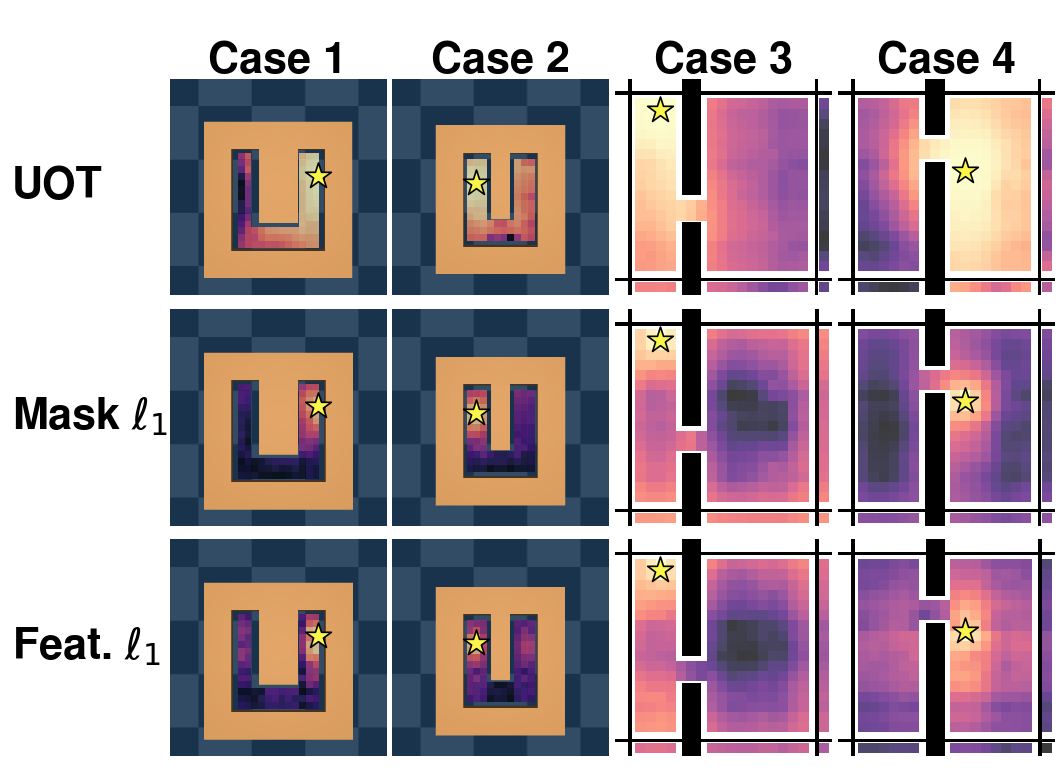}
\vspace{-7mm}
\caption{Goal-loss landscapes in MazeR and TwoRoomR. Stars mark goals; lighter colors indicate lower loss.}
\label{fig:loss-landscapes}
\end{minipage}\hfill
\begin{minipage}[t]{0.58\linewidth}
\vspace{-5pt}
\centering
\begingroup
\sffamily\bfseries\scriptsize
\setlength{\tabcolsep}{1pt}
\begin{tabular}{@{}lccc@{}}
& Start & Goal 1 & Goal 2 \\
\raisebox{.055\linewidth}[0pt][0pt]{External 1} & \includegraphics[width=.245\linewidth]{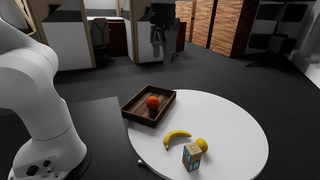} & \includegraphics[width=.245\linewidth]{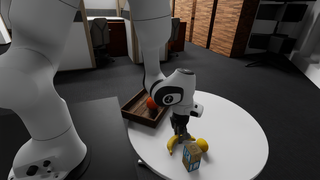} & \includegraphics[width=.245\linewidth]{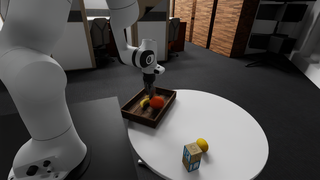} \\
\raisebox{.055\linewidth}[0pt][0pt]{External 2} & \includegraphics[width=.245\linewidth]{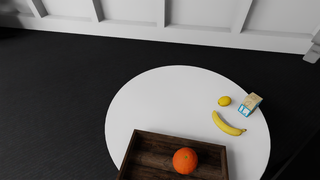} & \includegraphics[width=.245\linewidth]{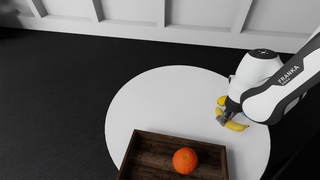} & \includegraphics[width=.245\linewidth]{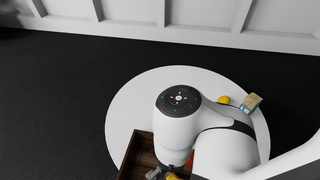} \\
\raisebox{.055\linewidth}[0pt][0pt]{Wrist} & \includegraphics[width=.245\linewidth]{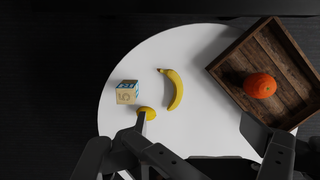} & \includegraphics[width=.245\linewidth]{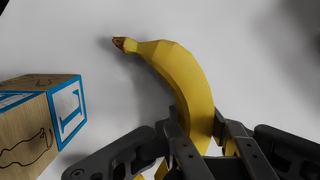} & \includegraphics[width=.245\linewidth]{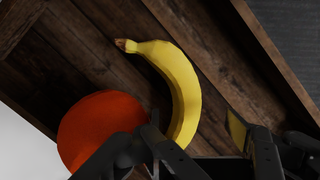} \\
\end{tabular}
\endgroup
\vspace{-1mm}
\caption{Three-view start and stage-wise goal images for a REALM-JEPA Place task.}
\label{fig:realm-goal-examples}
\end{minipage}
\end{figure}

\paragraph{Main results.} \Tabref{tab:planar-results} reports success rates and indicates whether goal-state input is used. We draw two conclusions:
\textbf{1) Slot-JEPA improves visual planning in both ID and OOD environments.} Using only a visual goal, it succeeds in 100\%, 100\%, and 92\% of ID cases and 90\%, 96\%, and 58\% of OOD cases. It also outperforms every baseline that receives the low-dimensional goal state in all four environments where goal-state results are available.
\textbf{2) Slot masks and UOT cost provide a planning signal aligned with progress.} As shown in \Figref{fig:loss-landscapes}, the obstacle-aware UOT loss changes gradually even far from the goal, which provides a smoother landscape compared to $\ell_1$ losses. Stable slots and object-level mask decomposition provide the foundation for applying UOT for planning. Additional loss landscapes and predicted-mask rollouts are collected in Appendix~\ref{sec:appendix-visualizations}.

\begin{table}[!tp]
\caption{Robotic arm planning success rates on REALM-JEPA (count and \%).}
\label{tab:robot-results}
\centering
\small
\resizebox{1.0\linewidth}{!}{%
\begin{tabular}{lccccc}
\toprule
Method & Reach & Pick & Rotate & Place & Overall \\
\midrule
V-JEPA 2.1 & \heatcell{taskseg}{15}{7.2em}{13/18 (72.2)} & \heatcell{taskdepth}{15}{7.2em}{19/30 (63.3)} & \heatcell{taskvos}{15}{7.2em}{6/30 (20.0)} & \heatcell{taskplace}{0}{7.2em}{0/30 (0.0)} & \heatcell{taskoverall}{15}{8.2em}{38/108 (35.2)} \\
V-JEPA 2.1 + TMP & \heatcell{taskseg}{45}{7.2em}{16/18 (88.9)} & \heatcell{taskdepth}{34}{7.2em}{22/30 (73.3)} & \heatcell{taskvos}{33}{7.2em}{13/30 (43.3)} & \heatcell{taskplace}{15}{7.2em}{12/30 (40.0)} & \heatcell{taskoverall}{38}{8.2em}{63/108 (58.3)} \\
\textbf{Slot-JEPA} & \heatcell{taskseg}{65}{7.2em}{\textbf{18/18 (100.0)}} & \heatcell{taskdepth}{65}{7.2em}{\textbf{27/30 (90.0)}} & \heatcell{taskvos}{65}{7.2em}{\textbf{25/30 (83.3)}} & \heatcell{taskplace}{65}{7.2em}{\textbf{23/30 (76.7)}} & \heatcell{taskoverall}{65}{8.2em}{\textbf{93/108 (86.1)}} \\
\bottomrule
\end{tabular}}
\end{table}

\begin{table}[!tp]
\caption{Spearman correlations among observed/predicted loss and EE--goal distance.}
\label{tab:robot-correlation}
\centering
\small
\setlength{\tabcolsep}{3pt}
\resizebox{1.0\linewidth}{!}{%
\begin{tabular}{@{}lccccccccc@{}}
\toprule
\raisebox{-7pt}[0pt][0pt]{Method} & \multicolumn{3}{c}{Obs. loss vs. EE--goal dist.}
& \multicolumn{3}{c}{Pred. loss vs. EE--goal dist.}
& \multicolumn{3}{c}{Pred. loss vs. Obs. loss} \\
\cmidrule(lr){2-4}\cmidrule(lr){5-7}\cmidrule(l){8-10}
& Ext. 1 & Ext. 2 & Wrist & Ext. 1 & Ext. 2 & Wrist & Ext. 1 & Ext. 2 & Wrist \\
\midrule
V-JEPA 2.1 & \heatcell{taskseg}{30}{4.2em}{0.350} & \heatcell{taskseg}{34}{4.2em}{0.407} & \heatcell{taskseg}{15}{4.2em}{0.533} & \heatcell{taskdepth}{40}{4.2em}{0.503} & \heatcell{taskdepth}{29}{4.2em}{0.453} & \heatcell{taskdepth}{15}{4.2em}{0.297} & \heatcell{taskvos}{35}{4.2em}{0.511} & \heatcell{taskvos}{38}{4.2em}{0.609} & \heatcell{taskvos}{15}{4.2em}{0.549} \\
V-JEPA 2.1 + TMP & \heatcell{taskseg}{15}{4.2em}{0.154} & \heatcell{taskseg}{15}{4.2em}{0.103} & \heatcell{taskseg}{37}{4.2em}{0.652} & \heatcell{taskdepth}{15}{4.2em}{0.327} & \heatcell{taskdepth}{15}{4.2em}{0.319} & \heatcell{taskdepth}{30}{4.2em}{0.454} & \heatcell{taskvos}{15}{4.2em}{0.409} & \heatcell{taskvos}{15}{4.2em}{0.421} & \heatcell{taskvos}{42}{4.2em}{0.720} \\
\textbf{Slot-JEPA} & \heatcell{taskseg}{65}{4.2em}{\textbf{0.792}} & \heatcell{taskseg}{65}{4.2em}{\textbf{0.891}} & \heatcell{taskseg}{65}{4.2em}{\textbf{0.807}} & \heatcell{taskdepth}{65}{4.2em}{\textbf{0.677}} & \heatcell{taskdepth}{65}{4.2em}{\textbf{0.788}} & \heatcell{taskdepth}{65}{4.2em}{\textbf{0.814}} & \heatcell{taskvos}{65}{4.2em}{\textbf{0.668}} & \heatcell{taskvos}{65}{4.2em}{\textbf{0.825}} & \heatcell{taskvos}{65}{4.2em}{\textbf{0.865}} \\
\bottomrule
\end{tabular}}
\end{table}

\begin{figure}[!tp]
\centering
\includegraphics[width=0.85\linewidth]{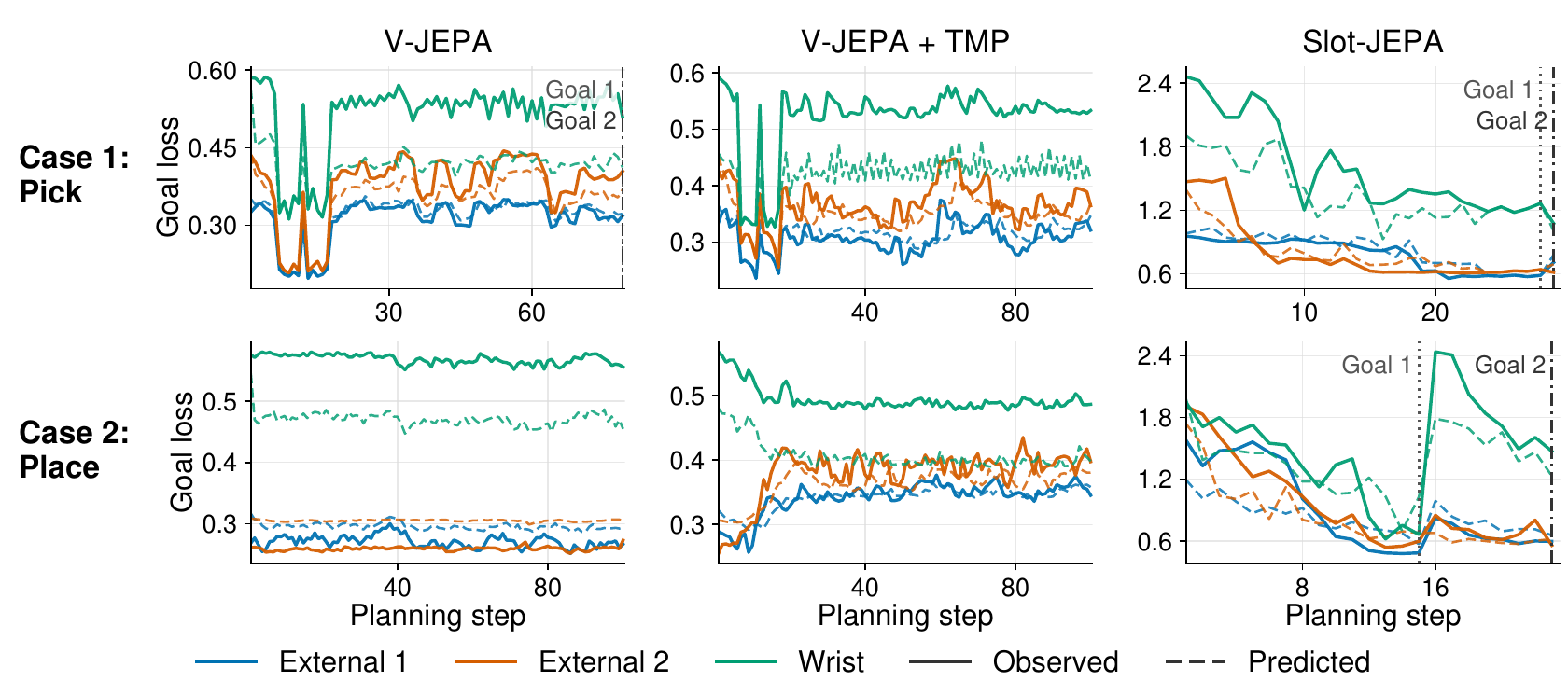}
\vspace{-3mm}
\caption{Goal-loss trajectories for Pick and Place. Panels use independent scales.}
\label{fig:realm-loss-curves}
\end{figure}

\subsection{Robotic Arm Planning}
\label{sec:robot}

\paragraph{Dataset \& evaluation protocol.}
Based on the REALM manipulation environment and task configurations \cite{sedlacek2025realm}, we construct the \textbf{REALM-JEPA} dataset for JEPA-based visual-goal planning. It provides start observations and stage-wise visual goals for 108 cases in four task families: 18 \textit{Reach} cases and 30 each for \textit{Pick}, \textit{Rotate}, and \textit{Place}. We fine-tune the robot dynamics predictor on DROID trajectories \cite{khazatsky2024droid}. The training settings are detailed in Appendix~\ref{sec:appendix-robot}.

In each task, we control a 7-DoF robot arm: six dimensions describe the end-effector (EE) pose, and one specifies the gripper opening angle. At step $t$, the visual observation $v_t$ contains images from two fixed external cameras and one wrist camera. Following V-JEPA 2 \cite{assran2025vjepa2}, we provide up to two stage-wise visual goals, each comprising images from the same three views, as illustrated in \Figref{fig:realm-goal-examples}. The low-dimensional state $r_t$ contains the current EE pose, gripper angle, and camera poses. The action $a_t$ specifies changes to the EE pose and gripper angle. The three views share one predictor, but maintain separate slots, transition memories, and predictions. At each MPC step, the two fixed views first update the CEM distribution; the wrist view then starts from that distribution, continues the search, and selects the action for execution. An inverse-kinematics controller converts the selected absolute EE pose into joint targets, which the arm tracks using PD feedback. Each planning attempt is limited to 100 MPC steps. Since $\ell_1$ losses provide weak guidance at large displacements, we use ground-truth target-position guidance for the patch-based baselines until the EE is within 30\,cm of the target position. Slot-JEPA uses visual goal only throughout. Full multi-view and guidance details are provided in Appendix~\ref{sec:appendix-robot}.

\paragraph{Quantitative results.}
\Tabref{tab:robot-results} reports success rates for the four task families. To assess whether the visual goal loss reflects physical progress, \Tabref{tab:robot-correlation} reports its Spearman correlation with 7-DoF EE--goal distance, using both losses predicted by the predictor and losses computed from the observations after execution. We also examine how the observed goal loss changes over planning steps in \Figref{fig:realm-loss-curves}. Appendix~\ref{sec:appendix-robot} gives the complete correlation definitions and statistics.
These results support three conclusions:
\textbf{1) Slot-JEPA improves success across all four task families, especially Rotate and Place.} Its overall success rate is 86.1\%, compared with 35.2\% for V-JEPA 2.1 and 58.3\% for V-JEPA 2.1 + TMP. On Rotate and Place, Slot-JEPA reaches 83.3\% and 76.7\%, respectively, compared with 20.0\% and 0.0\% for V-JEPA 2.1.
\textbf{2) Pointwise $\ell_1$ losses give weak guidance far from the goal.} Despite this ground-truth guidance, both patch-based baselines still fail often. Their losses computed from actual observations also correlate less strongly with EE--goal distance than Slot-JEPA's loss, indicating that predictor error alone cannot explain these failures.
\textbf{3) Replacing V-JEPA 2.1's predictor with TMP substantially improves planning.} The success rate rises from 35.2\% to 58.3\% while retaining the patch representation and goal loss. This gain is consistent with TMP retrieving useful information from the full trajectory history, whereas the original predictor relies on temporally contiguous frames and is restricted to nearby observations.

\begin{figure}[!tbp]
\centering
\begin{minipage}[t]{0.47\linewidth}
\vspace{0pt}
\centering
\noindent\vbox{\offinterlineskip
\hbox{\includegraphics[width=\linewidth]{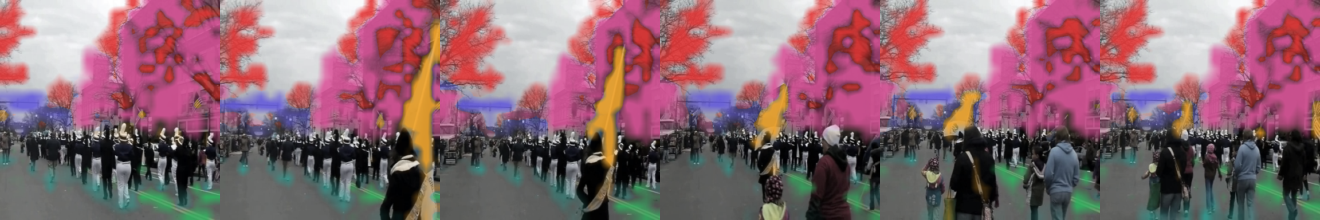}}
\kern2pt
\hbox{\includegraphics[width=\linewidth]{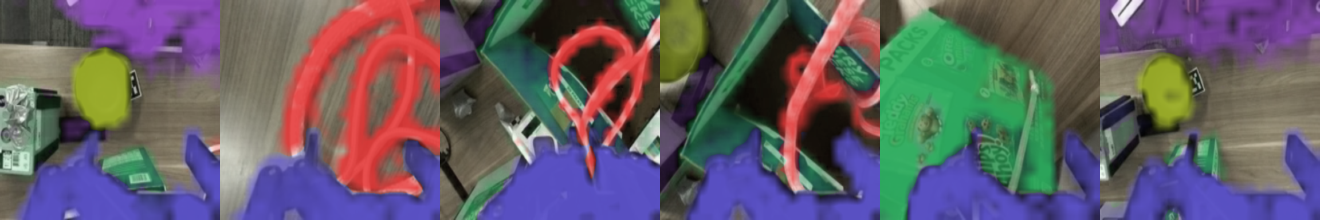}}
\kern2pt
\hbox{\includegraphics[width=\linewidth]{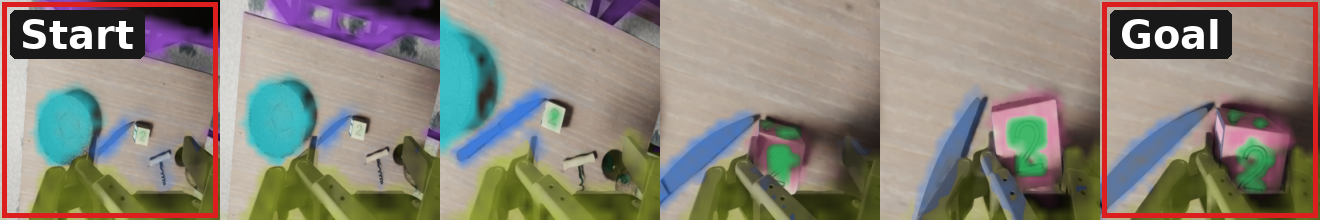}}
}\par
\caption{Slot-mask overlays from Kinetics-400, DROID, and REALM-JEPA (top to bottom). Colors track slots across each sequence.}
\label{fig:slot-overlays}
\end{minipage}\hfill
\begin{minipage}[t]{0.48\linewidth}
\vspace{0pt}
\centering
\includegraphics[width=\linewidth]{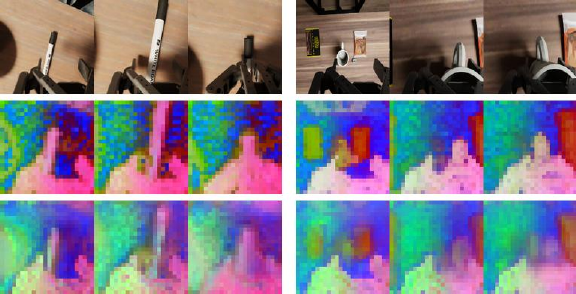}
\caption{Wrist-view RGB, observed features, and reconstructed features from predicted masks without seeing the frame (top to bottom).}
\label{fig:realm-observed-predicted}
\end{minipage}
\end{figure}

\paragraph{Visualizations.}
\Figref{fig:slot-overlays} shows examples from Kinetics-400 pretraining, DROID downstream training, and REALM-JEPA planning inference, respectively. The selected slots remain temporally consistent despite object motion, camera viewpoint changes, and objects leaving the view. Although the slot decomposition is frozen after pretraining, it continues to produce coherent masks in the unseen downstream scenes. Predicted future representations also closely match those extracted from subsequent observations (\Figref{fig:realm-observed-predicted}), supporting TMP's ability to model action-conditioned dynamics. Full-scale slot decompositions, additional UOT landscapes, and predicted-mask rollouts are provided in Appendix~\ref{sec:appendix-visualizations}.

\FloatBarrier
\subsection{Ablation Studies}
\label{sec:ablation}

\Tabref{tab:ablation} compares five configurations. \textbf{Line 1}: replace TMP with a V-JEPA 2.1-style sequential predictor \cite{murlabadia2026vjepa21} using recent transitions while retaining the mask interface and two-branch design. \textbf{Line 2}: remove UOT from planning. \textbf{Line 3}: give TMP only recent transitions. \textbf{Line 4}: use random transitions from the memory bank. \textbf{Line 5}: use the full Slot-JEPA model with transitions sampled according to \Eqref{eq:context-sampling}.

\begin{table}[!tp]
\caption{Ablation success rates (\%) across six planar environments and robot Reach. Bold marks the highest value in each column.}
\label{tab:ablation}
\centering
\setlength{\tabcolsep}{2.5pt}
\footnotesize
\resizebox{\linewidth}{!}{%
\begin{tabular}{@{}clllccccccc@{}}
\toprule
ID & Predictor & Planning loss & Context & TwoRoom & Maze & PushT & TwoRoomR & MazeR & PushObj & Reach \\
\midrule
1 & Sequential & UOT + mask $\ell_1$ & Recent transitions & \tworoomcell{65}{\textbf{100}} & \mazecell{65}{\textbf{100}} & \pushcell{15}{76} & \tworoomcell{65}{\textbf{94}} & \mazecell{65}{\textbf{98}} & \pushcell{15}{40} & -- \\
2 & TMP & Mask $\ell_1$ & Sampled via \Eqref{eq:context-sampling} & \tworoomcell{32}{96} & \mazecell{36}{92} & \pushcell{34}{82} & \tworoomcell{50}{78} & \mazecell{50}{90} & \pushcell{43}{50} & \heatcell{taskplace}{52}{3.8em}{88.9} \\
3 & TMP & UOT + mask $\ell_1$ & Recent transitions & \tworoomcell{15}{94} & \mazecell{15}{86} & \pushcell{15}{76} & \tworoomcell{56}{84} & \mazecell{30}{80} & \pushcell{65}{\textbf{58}} & \heatcell{taskplace}{15}{3.8em}{55.6} \\
4 & TMP & UOT + mask $\ell_1$ & Random transitions & \tworoomcell{32}{96} & \mazecell{22}{88} & \pushcell{28}{80} & \tworoomcell{15}{40} & \mazecell{15}{72} & \pushcell{37}{48} & \heatcell{taskplace}{59}{3.8em}{94.4} \\
\textbf{5} & TMP & UOT + mask $\ell_1$ & Sampled via \Eqref{eq:context-sampling} & \tworoomcell{65}{\textbf{100}} & \mazecell{65}{\textbf{100}} & \pushcell{65}{\textbf{92}} & \tworoomcell{61}{90} & \mazecell{61}{96} & \pushcell{65}{\textbf{58}} & \heatcell{taskplace}{65}{3.8em}{\textbf{100.0}} \\
\bottomrule
\end{tabular}}
\end{table}

These comparisons support three conclusions:
\textbf{1) UOT is important for planning.} Comparing lines 2 and 5, UOT improves success in every planar environment and robot Reach, even with goal-pose guidance for line 2.
\textbf{2) Complex planar tasks benefit from TMP's diverse transition context.} Lines 1 and 5 perform comparably on TwoRoom, TwoRoomR, Maze, and MazeR, consistent with recent information being most useful for these simpler tasks. On PushT and PushObj, line 5 performs better, highlighting the value of TMP's access to diverse historical transitions.
\textbf{3) Context must balance relevance and diversity.} Lines 3 and 5 show the benefit of drawing from a broader history; lines 4 and 5 show that query-conditioned relevance matters, especially under layout shift.

\section{Limitations}
\label{sec:limitations}

Slot decomposition is trained without instance or part annotations, using feature reconstruction rather than physical identity as its objective. Its masks support planning, but one slot may merge multiple objects or split one object, so consistent instance- or part-level decomposition is not guaranteed. This can limit reasoning about specific components and contact points. TMP builds memory from observed transitions and retrieves context using fixed state and mask cues. These cues may miss contact or appearance changes and select redundant examples. Better memory construction and retrieval criteria that balance relevance with diversity remain open problems.

\section{Conclusion}
\label{sec:conclusion}

We introduced Slot-JEPA, which factorizes dense visual representations into stable object-structured slots and time-varying spatial masks for action-conditioned prediction. This structure enables a UOT planning loss that is better aligned with spatial progress when pointwise losses provide weak guidance. TMP further draws on relevant transitions beyond a short contiguous history. Across seven visual-understanding benchmarks, mixtures of encoder and slot-reconstructed features remain transferable, while six planar environments and four Robotic Arm Planning tasks in the REALM-JEPA dataset show improved planning success, especially on the more complex manipulation tasks. These results show the value of combining structured mask prediction, spatially informed goal costs, and transition memory for visual planning.

\bibliography{references}
\bibliographystyle{iclr2027_conference}

\clearpage
\appendix
\phantomsection
\pdfbookmark[1]{Appendix Contents}{appendix-contents}
\section*{Appendix Contents}
\vspace{-0.8ex}
\noindent\rule{\linewidth}{0.4pt}
\vspace{0.8ex}
\begingroup
\setlength{\parindent}{0pt}
\setlength{\parskip}{1pt}
\newcommand{\appcontentsline}[3]{%
  \par\addvspace{0.6ex}%
  \noindent\makebox[2.5em][l]{\textbf{#2}}%
  \textbf{\hyperref[#1]{\textcolor{linkblue}{#3}}}%
  \hfill\textbf{\hyperref[#1]{\textcolor{black}{\pageref*{#1}}}}\par}
\newcommand{\appcontentssubline}[3]{%
  \noindent\hspace*{1.5em}\makebox[2.8em][l]{#2}%
  \hyperref[#1]{\textcolor{linkblue}{#3}}%
  \dotfill\hyperref[#1]{\textcolor{black}{\pageref*{#1}}}\par}
\appcontentsline{sec:appendix-method}{A}{Additional Method Details}
\appcontentssubline{sec:appendix-slot-config}{A.1}{Complete Slot Decomposition}
\appcontentssubline{sec:appendix-memory-construction}{A.2}{Predictor Architecture, Pretraining, and Memory}
\appcontentssubline{sec:appendix-uot-computation}{A.3}{UOT and Static-Slot Identification}
\appcontentsline{sec:appendix-experiments}{B}{Experimental Details and Complete Results}
\appcontentssubline{sec:appendix-pretraining}{B.1}{Pretraining}
\appcontentssubline{sec:appendix-visual}{B.2}{Visual Understanding}
\appcontentssubline{sec:appendix-planar}{B.3}{Planar Planning}
\appcontentssubline{sec:appendix-robot}{B.4}{Robotic Arm Planning}
\appcontentssubline{sec:appendix-visualizations}{B.5}{Additional Visualizations}
\endgroup
\clearpage
\section{Additional Method Details}
\label{sec:appendix-method}

This appendix gives the architecture and procedures shared by pretraining and all downstream tasks. Environment-specific slot and memory initialization, UOT settings, and planning procedures are reported in Appendix~\ref{sec:appendix-experiments}.

\subsection{Complete Slot Decomposition}
\label{sec:appendix-slot-config}

\paragraph{Reference features and initialization.}
During pretraining, the reference set $\mathcal{V}^{S}$ consists only of the observed/context frames in the sampled clip; none of the target frames to be predicted are used to construct the slots. We encode these frames once and partition their patch features into non-overlapping windows spanning $4\times4$ spatial patches. Within each window, four input patch tokens at evenly spaced valid positions initialize four local slots. This feature-dependent initialization avoids introducing a separate learned slot template while retaining spatial diversity.

\begin{figure}[!t]
\centering
\resizebox{0.9\linewidth}{!}{\begin{tikzpicture}[
  font=\small, >=Latex,
  flow/.style={-{Latex[length=1.7mm]}, line width=.65pt, draw=black!75},
  skip/.style={-{Latex[length=1.5mm]}, line width=.6pt, draw=black!65, rounded corners=2pt},
  link/.style={densely dotted, line width=.75pt, draw=black!55},
  mainpanel/.style={draw=black!38, fill=black!3, rounded corners=5pt,
                    line width=.7pt, minimum width=94mm},
  detailpanel/.style={draw=black!45, fill=black!1, rounded corners=5pt,
                      densely dotted, line width=.75pt, minimum width=68mm},
  io/.style={draw=taskplace!85!black, fill=taskplace!10, rounded corners=2pt,
             minimum width=34mm, minimum height=6.2mm, align=center, inner sep=1.7pt},
  enc/.style={draw=taskdepth!90!black, fill=taskdepth!13, rounded corners=2pt,
              minimum width=48mm, minimum height=6.2mm, align=center, inner sep=1.7pt},
  dec/.style={draw=taskseg!88!black, fill=taskseg!13, rounded corners=2pt,
              minimum width=48mm, minimum height=6.2mm, align=center, inner sep=1.7pt},
  head/.style={draw=taskvos!88!black, fill=taskvos!12, rounded corners=2pt,
               minimum width=48mm, minimum height=6.2mm, align=center, inner sep=1.7pt},
  cell/.style={draw=black!65, fill=white, rounded corners=1.5pt,
               minimum width=27mm, minimum height=5.5mm, align=center,
               inner sep=1.4pt, font=\scriptsize},
  add/.style={circle, draw=black!70, fill=white, minimum size=4mm,
              inner sep=0pt, font=\scriptsize},
  repeat/.style={font=\scriptsize, text=black!65, anchor=west},
  maintitle/.style={font=\bfseries\large, anchor=west},
  detailtitle/.style={font=\bfseries, anchor=west}
]

\node[maintitle, anchor=center, text=taskdepth!75!black] (etitle) at (5.00,11.75) {Two-stage Slot Encoder};
\node[io] (ein) at (5.00,4.95) {reference features $\bm Z^S$\\[-2pt]\scriptsize $B\times T\times H\times W\times D$};
\node[enc] (window) at (5.00,5.90) {window partition: $2\times4\times4$};
\node[enc] (local) at (5.00,6.87) {Local Slot Attention\\[-2pt]\scriptsize $4$ slots/window, width $D$};
\node[repeat] at (7.55,6.87) {$\times3$ iterations};
\node[enc] (pool) at (5.00,7.86) {concatenate local slots: $n_w\times4\times D$};
\node[enc] (fps) at (5.00,8.85) {farthest-point sampling: initialize $K=16$};
\node[dec] (global) at (5.00,9.84) {Global Slot Attention\\[-2pt]\scriptsize inputs: pooled local-slot candidates};
\node[repeat] at (7.55,9.84) {$\times3$ iterations};
\node[io, draw=taskseg!85!black, fill=taskseg!13] (eslots) at (5.00,10.78) {shared slot bank $\bm S$: $B\times K\times D$};
\draw[flow] (ein) -- (window);
\draw[flow] (window) -- (local);
\draw[flow] (local) -- (pool);
\draw[flow] (pool) -- (fps);
\draw[flow] (fps) -- (global);
\draw[flow] (global) -- (eslots);
\coordinate (encframetop) at (5.00,12.05);
\coordinate (encframebottom) at (5.00,3.90);
\begin{scope}[on background layer]
  \node[mainpanel, fit=(encframetop)(encframebottom)(etitle)(ein)(eslots), inner ysep=2.5mm] (encoderpanel) {};
\end{scope}

\node[detailtitle, anchor=center] (stitle) at (13.85,11.75) {Slot Attention Cell};
\node[io, minimum width=26mm] (iinput) at (12.25,4.25) {input tokens};
\node[io, minimum width=26mm] (sinput) at (15.45,4.25) {previous slots};
\node[cell] (iln) at (12.25,4.95) {LayerNorm};
\node[cell] (sln) at (15.45,4.95) {LayerNorm};
\node[cell] (kv) at (12.25,5.65) {Linear $K,V$};
\node[cell] (q) at (15.45,5.65) {Linear $Q$};
\node[cell, minimum width=37mm] (attn) at (13.85,6.45) {$A=\operatorname{softmax}_{\rm slot}(KQ^\top/\sqrt D)$};
\node[cell, minimum width=37mm] (norm) at (13.85,7.25) {normalize $A$ over input tokens};
\node[cell, minimum width=37mm] (weighted) at (13.85,8.05) {$U=A^\top V$};
\node[cell, minimum width=37mm] (gru) at (13.85,8.85) {GRU$(U,S_{\rm prev})$};
\node[cell, minimum width=37mm] (mlp) at (13.85,9.65) {LayerNorm $\rightarrow$ MLP};
\node[add] (sadd) at (13.85,10.35) {$+$};
\node[io, minimum width=37mm] (sout) at (13.85,10.95) {updated slots};
\draw[flow] (iinput) -- (iln);
\draw[flow] (sinput) -- (sln);
\draw[flow] (iln) -- (kv);
\draw[flow] (sln) -- (q);
\draw[flow] (kv.north) -- ([xshift=-9mm]attn.south);
\draw[flow] (q.north) -- ([xshift=9mm]attn.south);
\draw[flow] (attn) -- (norm);
\draw[flow] (norm) -- (weighted);
\draw[flow] (weighted) -- (gru);
\draw[flow] (gru) -- (mlp);
\draw[flow] (mlp) -- (sadd);
\draw[skip] (gru.west) -- ++(-.42,0) |- (sadd.west);
\draw[flow] (sadd) -- (sout);
\coordinate (slotframetop) at (13.85,12.05);
\coordinate (slotframebottom) at (13.85,3.90);
\coordinate (slotframeleft) at (10.55,8.00);
\coordinate (slotframeright) at (17.15,8.00);
\begin{scope}[on background layer]
  \node[detailpanel, fit=(slotframetop)(slotframebottom)(slotframeleft)(slotframeright)(stitle)(iinput)(sinput)(sout), inner ysep=2.5mm] (slotpanel) {};
\end{scope}
\coordinate (locallink) at ([yshift=-3mm]local.east);
\coordinate (globallink) at ([yshift=3mm]global.east);
\draw[link] (local.east) -- (locallink) -- (locallink -| slotpanel.west);
\draw[link] (global.east) -- (globallink) -- (globallink -| slotpanel.west);

\begin{scope}[yshift=-4mm]
\node[maintitle, anchor=center, text=taskvos!75!black] (dtitle) at (5.00,3.37) {Spatial Mask Decoder};
\node[io, minimum width=31mm] (dslots) at (3.00,-3.70) {shared slots $\bm S$\\[-2pt]\scriptsize $B\times K\times D$};
\node[io, minimum width=31mm] (patch) at (7.00,-3.70) {patch tokens $\bm Z$\\[-2pt]\scriptsize $BT\times N\times D$};
\node[dec] (dmain) at (7.00,-2.60) {Decoder Block $\times2$\\[-2pt]\scriptsize slot cross-attn $+$ patch encoder};
\node[enc, minimum width=35mm] (sprj) at (3.00,-1.50) {slot projection $D\!\rightarrow\!128$};
\node[enc, minimum width=35mm] (pprj) at (7.00,-1.50) {patch projection $D\!\rightarrow\!128$};
\node[head] (concat) at (5.00,-.40) {broadcast $+$ concatenate each patch--slot pair ($256$-d)};
\node[head] (maskhead) at (5.00,.60) {Mask MLP: $256\!\rightarrow\!128\!\rightarrow\!128\!\rightarrow\!1$};
\node[head] (dsoft) at (5.00,1.57) {softmax over $K$ slots: $\bm M$};
\node[io, draw=taskvos!88!black, fill=taskvos!12] (drecon) at (5.00,2.49) {reconstruction: $\bar{\bm Z}_{t,n}=\sum_k m_{t,k,n}\bm s_k$};
\draw[flow] (patch) -- (dmain);
\draw[flow] (dslots.north |- dmain.west) -- (dmain.west);
\draw[flow] (dmain) -- (pprj);
\draw[flow] (dslots) -- (sprj);
\draw[flow] (sprj.north) -- ([xshift=-12mm]concat.south);
\draw[flow] (pprj.north) -- ([xshift=12mm]concat.south);
\draw[flow] (concat) -- (maskhead);
\draw[flow] (maskhead) -- (dsoft);
\draw[flow] (dsoft) -- (drecon);
\coordinate (decframetop) at (5.00,3.68);
\coordinate (decframebottom) at (5.00,-4.08);
\begin{scope}[on background layer]
  \node[mainpanel, fit=(decframetop)(decframebottom)(dtitle)(patch)(dslots)(drecon), inner ysep=2.5mm] (decoderpanel) {};
\end{scope}

\node[detailtitle, anchor=center] (dbtitle) at (13.85,3.37) {Decoder Block};
\node[io, minimum width=25mm] (xinput) at (12.95,-3.50) {patch sequence $X$};
\node[io, minimum width=20mm] (xslots) at (15.45,-3.50) {slots $S$};
\node[cell, minimum width=30mm] (xln) at (12.95,-2.50) {LayerNorm};
\node[cell, minimum width=38mm] (cross) at (13.85,-1.50) {8-head cross-attention\\[-2pt]\scriptsize $Q=X$; $K,V=S$};
\node[add] (xadd) at (13.85,-.50) {$+$};
\node[dec, minimum width=50mm, minimum height=14mm] (tenc) at (13.85,.90) {\textbf{TransformerEncoderLayer}\\[-1pt]\scriptsize patch self-attention $+$ residual $+$ LayerNorm\\[-1pt]\scriptsize FFN $(D\!\rightarrow\!4D\!\rightarrow\!D)$ $+$ residual $+$ LayerNorm};
\node[io, minimum width=38mm] (xout) at (13.85,2.43) {updated patch sequence};
\draw[flow] (xinput) -- (xln);
\draw[flow] (xln) -- (cross);
\draw[flow] (xslots.north) -- ([xshift=8mm]cross.south);
\draw[flow] (cross) -- (xadd);
\draw[skip] (xinput.west) -- ++(-.25,0) |- (xadd.west);
\draw[flow] (xadd) -- (tenc);
\draw[flow] (tenc) -- (xout);
\coordinate (blockframetop) at (13.85,3.68);
\coordinate (blockframebottom) at (13.85,-4.08);
\coordinate (blockframeleft) at (10.55,0.00);
\coordinate (blockframeright) at (17.15,0.00);
\begin{scope}[on background layer]
  \node[detailpanel, fit=(blockframetop)(blockframebottom)(blockframeleft)(blockframeright)(dbtitle)(xinput)(xslots)(xout), inner ysep=2.5mm] (blockpanel) {};
\end{scope}
\draw[link] (dmain.east) -- (blockpanel.west |- dmain.east);
\end{scope}

\end{tikzpicture}}
\caption{Network architecture of the slot decomposition module.}
\label{fig:appendix-slot-architecture}
\end{figure}
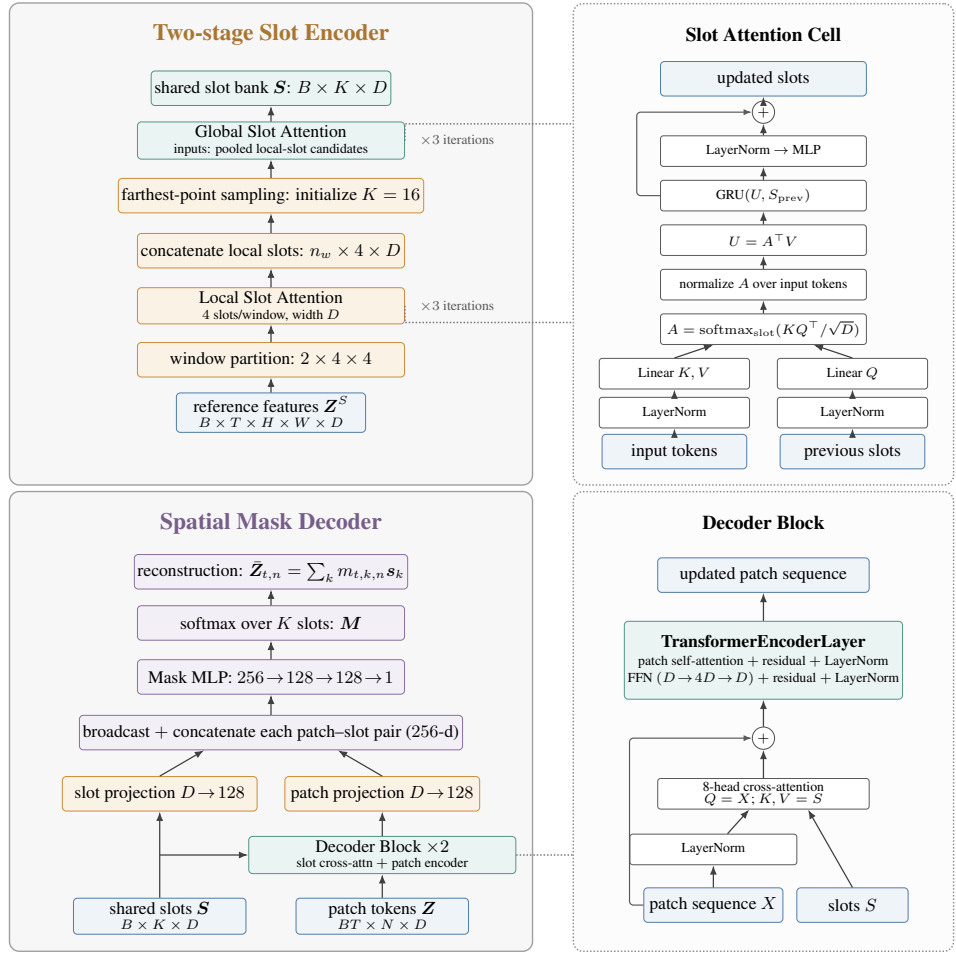

\paragraph{Local-to-global Slot Attention.}
Each local window is processed by three Slot Attention iterations. In every iteration, layer-normalized slots form queries and layer-normalized patch tokens form keys and values. Attention is normalized first across the competing slots and then across the input tokens assigned to each slot. The resulting weighted value is passed through a GRU update followed by a residual two-layer MLP, as formalized in \cref{sec:slot-decomposition}. We pool the local slots from all windows and select $K=16$ of them by farthest-point sampling (FPS) in feature space. FPS starts from a valid local slot and repeatedly chooses the slot with the largest distance to its nearest selected slot. The selected features initialize a second, global Slot Attention module, which applies the same three-step iterative update to the pooled local-slot candidates and produces the shared slot bank $\bm{S}$. The complete local-to-global extraction and mask-decoding network is shown in \Figref{fig:appendix-slot-architecture}.

\paragraph{Mask decoding and optimization.}
The mask decoder first projects patch features and slots to width 128. Two decoder blocks alternately update the patch queries by cross-attending to the fixed slot bank and by applying self-attention among patches. For every patch--slot pair, an MLP receives the concatenated decoded patch feature and slot feature and produces an assignment logit. A softmax across the $K$ slots yields $\bm{M}$, and the mask-weighted slot sum reconstructs the original dense representation as in \Eqref{eq:slot-reconstruction}. We jointly optimize the local and global Slot Attention modules and the mask decoder with the feature-reconstruction and slot-orthogonality terms in \cref{sec:slot-decomposition}. The same extracted slot bank is held fixed when decoding the observed, predicted, and goal frames of one environment. The environment-specific choice of inference reference frames and the slot-refresh policy are given in Appendix~\ref{sec:appendix-experiments}.

\subsection{Predictor Architecture, Pretraining, and Memory}
\label{sec:appendix-memory-construction}

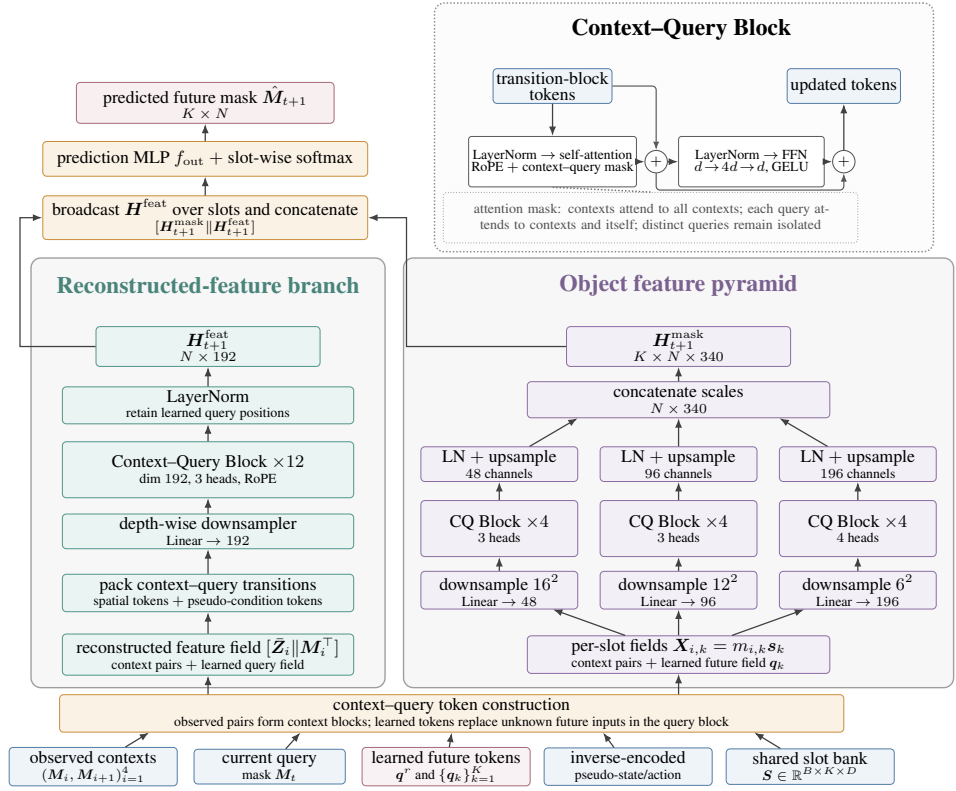
\begin{figure}[!t]
\centering
\resizebox{0.9\linewidth}{!}{\begin{tikzpicture}[
  font=\small, >=Latex,
  flow/.style={-{Latex[length=1.7mm]}, line width=.65pt, draw=black!75},
  skip/.style={-{Latex[length=1.5mm]}, line width=.6pt, draw=black!65, rounded corners=2pt},
  link/.style={densely dotted, line width=.75pt, draw=black!55},
  io/.style={draw=taskplace!85!black, fill=taskplace!10, rounded corners=2pt,
             minimum height=6.2mm, align=center, inner sep=1.7pt},
  global/.style={draw=taskseg!88!black, fill=taskseg!13, rounded corners=2pt,
                 minimum height=6.2mm, align=center, inner sep=1.7pt},
  object/.style={draw=taskvos!88!black, fill=taskvos!12, rounded corners=2pt,
                 minimum height=6.2mm, align=center, inner sep=1.7pt},
  fusion/.style={draw=taskdepth!90!black, fill=taskdepth!13, rounded corners=2pt,
                 minimum height=6.2mm, align=center, inner sep=1.7pt},
  cell/.style={draw=black!65, fill=white, rounded corners=1.5pt,
               minimum height=5.8mm, align=center, inner sep=1.4pt, font=\scriptsize},
  add/.style={circle, draw=black!70, fill=white, minimum size=4mm,
              inner sep=0pt, font=\scriptsize},
  mainpanel/.style={draw=black!38, fill=black!3, rounded corners=5pt,
                    line width=.7pt},
  detailpanel/.style={draw=black!45, fill=black!1, rounded corners=5pt,
                      densely dotted, line width=.75pt},
  title/.style={font=\bfseries\large, anchor=center},
  note/.style={font=\scriptsize, text=black!68, align=center}
]

\node[io, minimum width=30mm] (contexts) at (1.70,5.55)
  {observed contexts\\[-2pt]\scriptsize $(\bm M_i,\bm M_{i+1})_{i=1}^{4}$};
\node[io, minimum width=27mm] (querymask) at (4.85,5.55)
  {current query\\[-2pt]\scriptsize mask $\bm M_t$};
\node[io, minimum width=30mm, draw=taskoverall!85!black, fill=taskoverall!10]
  (futuretokens) at (8.00,5.55)
  {learned future tokens\\[-2pt]\scriptsize $\bm q^r$ and $\{\bm q_k\}_{k=1}^{K}$};
\node[io, minimum width=30mm] (pseudocond) at (11.25,5.55)
  {inverse-encoded\\[-2pt]\scriptsize pseudo-state/action};
\node[io, minimum width=30mm] (slots) at (14.50,5.55)
  {shared slot bank\\[-2pt]\scriptsize $\bm S\in\mathbb R^{B\times K\times D}$};

\node[fusion, minimum width=140mm] (assemble) at (8.10,6.50)
  {context--query token construction\\[-2pt]
   \scriptsize observed pairs form context blocks; learned tokens replace unknown future inputs in the query block};

\draw[flow] (contexts) -- ([xshift=-54mm]assemble.south);
\draw[flow] (querymask) -- ([xshift=-27mm]assemble.south);
\draw[flow] (futuretokens) -- (assemble.south);
\draw[flow] (pseudocond) -- ([xshift=27mm]assemble.south);
\draw[flow] (slots) -- ([xshift=54mm]assemble.south);

\node[title, text=taskseg!78!black] (gtitle) at (3.75,14.15)
  {Reconstructed-feature branch};
\node[global, minimum width=52mm] (gfield) at (3.75,7.55)
  {reconstructed feature field $[\bar{\bm Z}_i\Vert\bm M_i^\top]$\\[-2pt]
   \scriptsize context pairs $+$ learned query field};
\node[global, minimum width=52mm] (gpack) at (3.75,8.65)
  {pack context--query transitions\\[-2pt]
   \scriptsize spatial tokens $+$ pseudo-condition tokens};
\node[global, minimum width=52mm] (gproj) at (3.75,9.75)
  {depth-wise downsampler\\[-2pt]\scriptsize Linear $\rightarrow192$};
\node[global, minimum width=52mm, minimum height=10mm] (gtr) at (3.75,10.85)
  {Context--Query Block $\times12$\\[-2pt]
   \scriptsize dim $192$, $3$ heads, RoPE};
\node[global, minimum width=52mm] (gln) at (3.75,12.00)
  {LayerNorm\\[-2pt]\scriptsize retain learned query positions};
\node[io, minimum width=40mm, draw=taskseg!88!black, fill=taskseg!13] (hfeat) at (3.75,13.05)
  {$\bm H^{\rm feat}_{t+1}$\\[-2pt]\scriptsize $N\times192$};

\draw[flow] (assemble.north -| gfield.south) -- (gfield.south);
\draw[flow] (gfield) -- (gpack);
\draw[flow] (gpack) -- (gproj);
\draw[flow] (gproj) -- (gtr);
\draw[flow] (gtr) -- (gln);
\draw[flow] (gln) -- (hfeat);

\coordinate (gframeleft) at (.75,10.65);
\coordinate (gframeright) at (6.75,10.65);
\coordinate (gframetop) at (3.75,14.47);
\coordinate (gframebottom) at (3.75,7.12);
\begin{scope}[on background layer]
  \node[mainpanel, fit=(gframeleft)(gframeright)(gframetop)(gframebottom)
        (gtitle)(gfield)(hfeat), inner sep=1.6mm] (gbg) {};
\end{scope}

\node[title, text=taskvos!78!black] (otitle) at (12.15,14.15)
  {Object feature pyramid};
\node[object, minimum width=54mm] (ofield) at (12.15,7.55)
  {per-slot fields $\bm X_{i,k}=m_{i,k}\bm s_k$\\[-2pt]
   \scriptsize context pairs $+$ learned future field $\bm q_k$};

\node[object, minimum width=28mm] (s16in) at (8.95,8.70)
  {downsample $16^2$\\[-2pt]\scriptsize Linear $\rightarrow48$};
\node[object, minimum width=28mm] (s12in) at (12.15,8.70)
  {downsample $12^2$\\[-2pt]\scriptsize Linear $\rightarrow96$};
\node[object, minimum width=28mm] (s6in) at (15.35,8.70)
  {downsample $6^2$\\[-2pt]\scriptsize Linear $\rightarrow196$};

\node[object, minimum width=28mm, minimum height=10mm] (s16tr) at (8.95,9.80)
  {CQ Block $\times4$\\[-2pt]\scriptsize $3$ heads};
\node[object, minimum width=28mm, minimum height=10mm] (s12tr) at (12.15,9.80)
  {CQ Block $\times4$\\[-2pt]\scriptsize $3$ heads};
\node[object, minimum width=28mm, minimum height=10mm] (s6tr) at (15.35,9.80)
  {CQ Block $\times4$\\[-2pt]\scriptsize $4$ heads};

\node[object, minimum width=28mm] (s16out) at (8.95,10.95)
  {LN $+$ upsample\\[-2pt]\scriptsize $48$ channels};
\node[object, minimum width=28mm] (s12out) at (12.15,10.95)
  {LN $+$ upsample\\[-2pt]\scriptsize $96$ channels};
\node[object, minimum width=28mm] (s6out) at (15.35,10.95)
  {LN $+$ upsample\\[-2pt]\scriptsize $196$ channels};
\node[object, minimum width=54mm] (ocat) at (12.15,12.10)
  {concatenate scales\\[-2pt]\scriptsize $N\times340$};
\node[io, minimum width=40mm, draw=taskvos!88!black, fill=taskvos!12] (hmask) at (12.15,13.05)
  {$\bm H^{\rm mask}_{t+1}$\\[-2pt]\scriptsize $K\times N\times340$};

\draw[flow] (assemble.north -| ofield.south) -- (ofield.south);
\draw[flow] (ofield) -- (s16in);
\draw[flow] (ofield) -- (s12in);
\draw[flow] (ofield) -- (s6in);
\draw[flow] (s16in) -- (s16tr);
\draw[flow] (s12in) -- (s12tr);
\draw[flow] (s6in) -- (s6tr);
\draw[flow] (s16tr) -- (s16out);
\draw[flow] (s12tr) -- (s12out);
\draw[flow] (s6tr) -- (s6out);
\draw[flow] (s16out) -- ([xshift=-18mm]ocat.south);
\draw[flow] (s12out) -- (ocat.south);
\draw[flow] (s6out) -- ([xshift=18mm]ocat.south);
\draw[flow] (ocat) -- (hmask);

\coordinate (oframeleft) at (7.40,10.65);
\coordinate (oframeright) at (16.90,10.65);
\coordinate (oframetop) at (12.15,14.47);
\coordinate (oframebottom) at (12.15,7.12);
\begin{scope}[on background layer]
  \node[mainpanel, fit=(oframeleft)(oframeright)(oframetop)(oframebottom)
        (otitle)(ofield)(s16in)(s6in)(hmask), inner sep=1.6mm] (obg) {};
\end{scope}

\node[fusion, minimum width=58mm] (fuse) at (3.70,15.35)
  {broadcast $\bm H^{\rm feat}$ over slots and concatenate\\[-2pt]
   \scriptsize $[\bm H^{\rm mask}_{t+1}\Vert\bm H^{\rm feat}_{t+1}]$};
\node[fusion, minimum width=58mm] (head) at (3.70,16.40)
  {prediction MLP $f_{\rm out}$ $+$ slot-wise softmax};
\node[io, minimum width=46mm, draw=taskoverall!85!black, fill=taskoverall!12] (out) at (3.70,17.40)
  {predicted future mask $\hat{\bm M}_{t+1}$\\[-2pt]\scriptsize $K\times N$};

\draw[flow] (hfeat.west) -- ++(-1.35,0) |- (fuse.west);
\draw[flow] (hmask.west) -- ++(-2.85,0) |- (fuse.east);
\draw[flow] (fuse) -- (head);
\draw[flow] (head) -- (out);

\node[title] (cqtitle) at (12.20,18.72) {Context--Query Block};

\node[io, minimum width=20mm] (xin) at (9.90,17.70)
  {transition-block\\[-2pt]tokens};
\node[io, minimum width=20mm] (xout) at (15.10,17.70)
  {updated tokens};

\node[cell, minimum width=29mm, minimum height=9mm] (mhsa) at (9.90,16.35)
  {LayerNorm $\rightarrow$ self-attention\\[-2pt]RoPE $+$ context--query mask};
\node[add] (a1) at (11.75,16.35) {$+$};
\node[cell, minimum width=26mm, minimum height=9mm] (ffn) at (13.45,16.35)
  {LayerNorm $\rightarrow$ FFN\\[-2pt]$d\!\rightarrow\!4d\!\rightarrow\!d$, GELU};
\node[add] (a2) at (15.10,16.35) {$+$};

\draw[flow] (xin.south) -- (mhsa.north);
\draw[flow] (mhsa.east) -- (a1.west);
\draw[flow] (a1.east) -- (ffn.west);
\draw[flow] (ffn.east) -- (a2.west);
\draw[flow] (a2.north) -- (xout.south);
\draw[skip] (xin.east) -- ++(.50,0) -| (a1.north);
\draw[skip] (a1.south) -- ++(0,-.30) -| (a2.south);

\node[note, draw=black!45, densely dotted, rounded corners=2pt,
      text width=70mm, inner sep=2mm] (masknote) at (11.65,15.35)
  {attention mask: contexts attend to all contexts; each query attends to contexts and itself; distinct queries remain isolated};
\draw[link] (masknote.north) -- (mhsa.south);

\coordinate (cqleft) at (8.45,16.70);
\coordinate (cqright) at (16.55,16.70);
\coordinate (cqtop) at (12.20,19.02);
\coordinate (cqbottom) at (12.20,15.05);
\begin{scope}[on background layer]
  \node[detailpanel, fit=(cqleft)(cqright)(cqtop)(cqbottom)
        (cqtitle)(xin)(xout)(masknote), inner sep=1.5mm] (cqpanel) {};
\end{scope}

\end{tikzpicture}}
\caption{TMP architecture. Observed transitions form the context blocks, while inverse-encoded pseudo-action and learned future tokens specify the query transition. The reconstructed-feature tower and the three-scale object feature pyramid run in parallel and are fused to predict the future slot mask.}
\label{fig:appendix-tmp-architecture}
\end{figure}

\paragraph{Predictor architecture.}
\Figref{fig:appendix-tmp-architecture} shows the layer-level architecture of the two TMP branches. TMP uses branch-specific context--query stacks. The reconstructed-feature branch has 12 blocks of width 192 with three attention heads. The mask branch applies separate object-wise predictors at three spatial scales, each with four blocks. Their $(\text{spatial size},\text{width},\text{heads})$ configurations are $(16,48,3)$, $(12,96,3)$, and $(6,196,4)$. The current and future fields are concatenated channel-wise at each scale; the unavailable future query field is replaced by learned tokens. After attention, we retain only the query positions associated with those future tokens, upsample the three outputs, and concatenate them to form $\bm{H}^{\mathrm{mask}}_{t+1}$. The shared feature branch uses the same context--query organization on the reconstructed feature field and masks to model interactions across slots. Its output is fused with every object-wise mask feature before the slot-wise softmax. These dimensions are fixed in comparisons with the sequential-predictor ablation unless stated otherwise. 

\paragraph{Bidirectional pretraining contexts.}
Each pretraining sample contains seven frames: four observed/context frames and three held-out target frames. We construct both directions from every clip. In the forward case, the first four frames predict the last three; in the backward case, the last four predict the first three. Thus the predictor learns a relation between observed transitions and a query transition without relying on a fixed chronological direction. The observed transitions from the same sampled clip directly form the training context, so pretraining neither retrieves from nor maintains a persistent memory bank.

\paragraph{Inverse condition encoders.}
Natural pretraining videos have no action labels. We therefore reconstruct each pair of source and target frames from their masks and shared slots, and feed the paired reconstructions and masks to an inverse transition encoder. A spatial projection, learned condition tokens, and an eight-layer, three-head Transformer of width 192 compress their change into the low-dimensional pseudo-action $\tilde a_t$ in \cref{sec:tmp}. A companion Transformer with the same dimensions maps each reconstructed frame and its masks to a pseudo-state. The target reaches TMP through these low-dimensional condition tokens rather than through the full target mask field. After pretraining, both inverse encoders are discarded and replaced by task-specific state and action encoders; the TMP body is transferred.

\paragraph{Memory construction and maintenance at inference.}
All entries in an inference memory bank are observed transitions decoded with one common slot bank. An entry stores $(\bm{M}_i,r_i,a_i,\bm{M}_{i+1},r_{i+1})$, and each newly executed transition is appended after its next observation becomes available. For a query, TMP computes relevance from normalized low-dimensional state, per-slot area, and per-slot centroid, then samples five transitions without replacement using \Eqref{eq:context-sampling}. One sampled context set is shared by all candidate actions in the same prediction request. When the slot bank is refreshed, retained frames are decoded again with the new slots and the associated memory masks are rebuilt, preserving slot-index consistency. The source of the initial nonempty bank and the refresh schedule depend on the environment and are specified in Appendix~\ref{sec:appendix-experiments}.

\subsection{UOT and Static-Slot Identification}
\label{sec:appendix-uot-computation}

\paragraph{UOT computation.}
The UOT term in \cref{sec:uot} is evaluated independently for corresponding slot masks. We retain each mask's original mass: the masks are neither divided by their areas nor constrained to have equal total mass. Instead, the relaxed marginal constraints allow mass to disappear or emerge at finite cost. For nonnegative vectors $x$ and $y$, we use the generalized KL divergence
\begin{equation}
\mathrm{KL}(x\,\|\,y)=\sum_i\left[x_i\log\frac{x_i}{y_i}-x_i+y_i\right].
\label{eq:generalized-kl}
\end{equation}
The default ground cost is $c_{ij}=\lVert\bm{\xi}_i-\bm{\xi}_j\rVert_2^2$ on normalized patch coordinates. Given $p,q\in\mathbb{R}_+^N$, we form the Gibbs kernel and unbalanced exponent
\begin{equation}
K_{ij}=\exp(-c_{ij}/\varepsilon),\qquad
\theta=\frac{\rho}{\rho+\varepsilon}.
\label{eq:uot-kernel}
\end{equation}
Algorithm~\ref{alg:generalized-sinkhorn} gives the implemented scaling procedure. All divisions and logarithms clamp their arguments below by $\delta=10^{-8}$ for numerical stability. The entropy parameter $\varepsilon$ enters through $K$ and the scaling exponent; after the final iteration, we evaluate the spatial transport and relaxed-marginal terms of \Eqref{eq:uot-objective} at the resulting plan.

\begin{algorithm}[ht]
\caption{Generalized Sinkhorn scaling for one aligned slot-mask pair}
\label{alg:generalized-sinkhorn}
\small
\begin{minipage}{0.96\linewidth}
\textbf{Input:} masks $p,q\in\mathbb{R}_+^N$, ground cost $C\in\mathbb{R}_+^{N\times N}$, entropy $\varepsilon$, mass penalty $\rho$, iteration count $L$, and numerical floor $\delta$.\\[-2pt]
\begin{enumerate}
\setlength{\itemsep}{1pt}
\setlength{\parskip}{0pt}
\setlength{\parsep}{0pt}
\item Set $a=\max(p,\delta)$, $b=\max(q,\delta)$, $K=\max(\exp(-C/\varepsilon),\delta)$, and $\theta=\rho/(\rho+\varepsilon)$ element-wise.
\item Initialize $u=\bm{1}_N$ and $v=\bm{1}_N$.
\item For $\ell=1,\ldots,L$, update
\begin{equation*}
u\leftarrow\left(\frac{a}{\max(Kv,\delta)}\right)^{\theta},\qquad
v\leftarrow\left(\frac{b}{\max(K^\top u,\delta)}\right)^{\theta}.
\end{equation*}
\item Construct $\pi=\operatorname{diag}(u)K\operatorname{diag}(v)$.
\end{enumerate}
\textbf{Return:}
$\langle\pi,C\rangle+\rho\,\mathrm{KL}(\pi\bm{1}\,\|\,a)+\rho\,\mathrm{KL}(\pi^\top\bm{1}\,\|\,b)$.
\end{minipage}
\end{algorithm}

To emphasize slots that must change between the reference mask $p^{\mathrm{ref}}$ and target mask $q$, we compute weights only from observed or target masks, never from the prediction:
\begin{equation}
\begin{aligned}
d_k&=\frac{1}{N}\lVert q_k-p^{\mathrm{ref}}_k\rVert_1,\quad
\bar d=\frac{1}{K}\sum_jd_j,\quad \\
w_k&=\operatorname{clip}\!\left(\frac{d_k}{\bar d},w_{\min},w_{\max}\right),\quad 
\mathcal{L}_{\mathrm{uot}}=\frac{\sum_kw_k\mathcal{T}_{\varepsilon,\rho}(p_k,q_k)}{\sum_kw_k}.
\end{aligned}
\label{eq:change-weighted-uot}
\end{equation}
If all $d_k$ vanish, we set all weights to one. We use $w_{\min}=0.05$ and $w_{\max}=5$; the same weights are applied to the mask-$\ell_1$ term. This reduction prevents persistent background slots from dominating the objective. 

\begin{figure}[!t]
\centering
\includegraphics[width=0.72\linewidth]{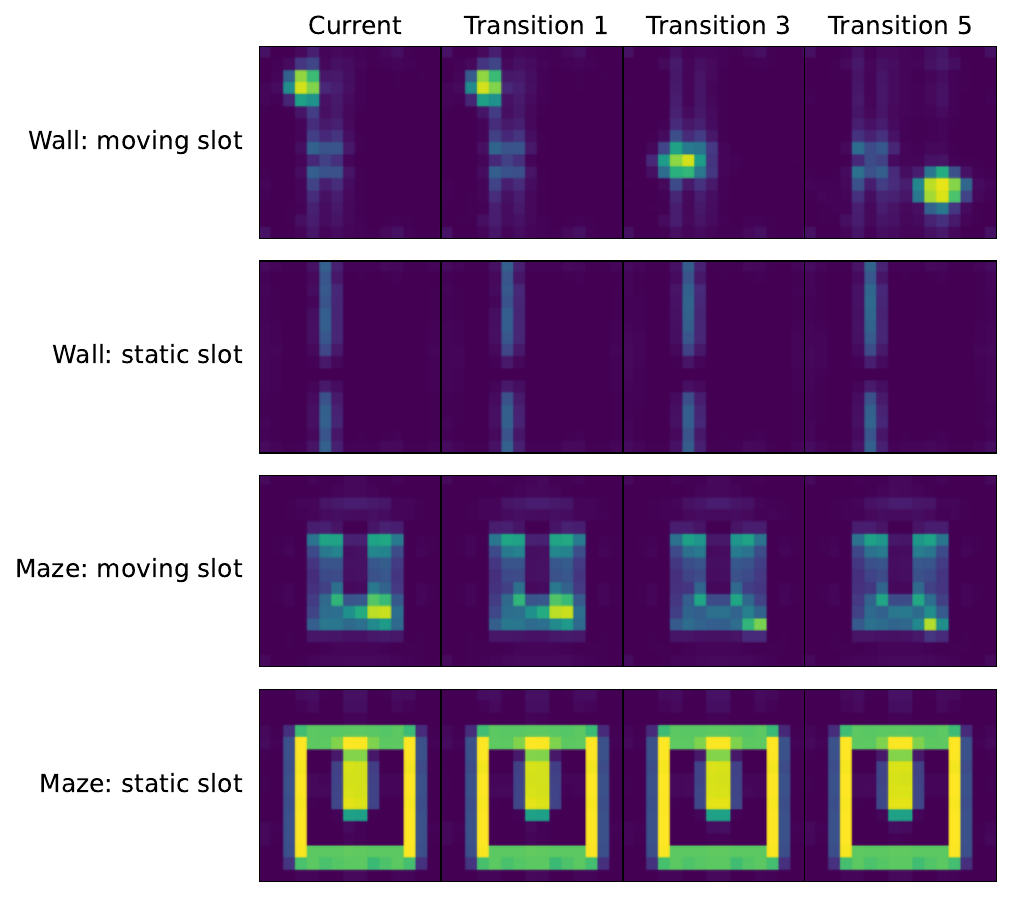}
\caption{Examples of slots identified from their predicted changes. Moving-slot masks follow the controlled entity across probe transitions, while static-slot masks remain aligned with the wall or maze geometry. The labels are produced by the change-score rule rather than ground-truth segmentation.}
\label{fig:static-slot-examples}
\end{figure}

\paragraph{Static-slot identification.}
We determine static slots from predicted mask changes rather than semantic labels. For each probe transition $p$ and slot $k$, we compare an aligned source mask $m_k$ with its model-predicted next mask $\hat m^{(p)}_k$. We retain the $L$ largest absolute patch-wise changes and average them. The model-level change score is
\begin{equation}
\Delta_k=\frac{1}{P}\sum_{p=1}^{P}\frac{1}{L}
\sum_{n\in\operatorname{TopL}(|\hat m^{(p)}_k-m_k|)}
|\hat m^{(p)}_{k,n}-m_{k,n}|.
\end{equation}
A slot is static when $\Delta_k\leq\tau_{\mathrm{static}}$ and moving when $\Delta_k>\tau_{\mathrm{moving}}$; slots in the gap are left unclassified. This two-threshold rule prevents borderline changes from contaminating either group. As illustrated in \Figref{fig:static-slot-examples}, moving slots shift with controlled entities, whereas static slots remain aligned with persistent scene geometry.

For obstacle-aware planning, we peak-normalize and optionally sharpen the selected static masks, then take their maximum to obtain an obstacle map. For a moving slot, this map modifies the ground cost as
\begin{equation}
c^{\mathrm{obs}}_{ij}=c_{ij}+\lambda_{\mathrm{obs}}r_{ij},
\end{equation}
where $r_{ij}$ measures obstacle occupancy along the path from patch $i$ to patch $j$. The ground cost therefore encodes obstacle geometry directly. The probe transitions, thresholds, and ground-cost variant are environment-specific and are reported with the corresponding experiments.

\section{Experimental Details and Complete Results}
\label{sec:appendix-experiments}

This appendix follows the four stages of the empirical study: representation pretraining, visual-understanding evaluation, planar planning, and robotic arm planning. We collect qualitative analyses in a final visualization subsection so that the representation, planning objective, and predicted rollouts can be compared in one place.

\subsection{Pretraining}
\label{sec:appendix-pretraining}

\paragraph{Initialization and data.}
We initialize the visual encoder from the released distilled V-JEPA 2.1 ViT-B/16 checkpoint \cite{murlabadia2026vjepa21}. This 80M-parameter encoder is distilled from the 2B-parameter V-JEPA 2.1 ViT-G teacher on VisionMix-163M, which combines LVD-142M images with videos from SSv2, Kinetics, HowTo100M, and YT-1B. We then pretrain Slot-JEPA on the training split of Kinetics-400 \cite{kay2017kinetics}, which contains roughly 240,000 clips. The complete dataset contains 306,245 clips from 400 human-action classes. Each clip lasts approximately 10 seconds and comes from a distinct YouTube video. Its content spans individual actions such as drawing and drinking, person--person interactions such as hugging and shaking hands, and person--object interactions such as opening objects and playing instruments. We decode the videos at $384\times384$ resolution.

\paragraph{Optimization.}
We use $K=16$ global slots and $4\times4$ spatial windows with four local slots per window. Each seven-frame clip supplies four observed frames and three target frames; the observed transitions provide context for predicting the targets. We jointly optimize the slot decomposition module, TMP, and the inverse transition encoders while fine-tuning the final six visual-encoder blocks; the remaining blocks are frozen. Pretraining runs for 10 epochs on eight RTX 4090 GPUs and takes about 30 hours. We use AdamW \cite{loshchilov2019adamw} with learning rate $10^{-4}$ and weight decay $10^{-2}$. The learning rate follows cosine decay \cite{loshchilov2017sgdr} to $10^{-5}$. We set $\lambda_{\mathrm{orth}}=0.001$ in $\mathcal{L}_{\mathrm{slot}}$ and $(\lambda_{\mathrm{uot}},\lambda_{\mathrm{mask}},\lambda_{\mathrm{feat}})=(3.0,3.0,0.1)$ in $\mathcal{L}_{\mathrm{pred}}$. UOT uses five Sinkhorn iterations and $\varepsilon=0.05$. For all downstream tasks, the pretrained encoder and slot decomposition module are frozen.

\subsection{Visual Understanding}
\label{sec:appendix-visual}

\paragraph{Downstream protocols.}
The encoder and decomposition module remain frozen for all three probes. Segmentation trains a linear head for 2,000 steps using Adam \cite{kingma2015adam} with batch size 256, selecting the learning rate from $\{10^{-4},3\!\times\!10^{-4},8\!\times\!10^{-4},10^{-3},2\!\times\!10^{-3},3\!\times\!10^{-3},10^{-2}\}$. Depth trains a linear scale-invariant logarithmic loss (SigLoss) head \cite{eigen2014depth} for 38,400 steps (AdamW, learning rate $3\times10^{-4}$, batch size 16). VOS requires no training: nearest-neighbor propagation uses the latest 20 frames, top-5 neighbors, and 16 frames to initialize slots from first-frame mask centers.

\paragraph{Competitors.}
We compare against MoCo v2 \cite{chen2020mocov2}, MoCo v3 \cite{chen2021mocov3}, SwAV \cite{caron2020swav}, DINO \cite{caron2021dino}, iBOT \cite{zhou2022ibot}, Mugs \cite{zhou2022mugs}, Leopart \cite{ziegler2022leopart}, Context Autoencoder (CAE) \cite{chen2024cae}, and Masked Autoencoders (MAE) \cite{he2022mae}. The locally evaluated V-JEPA 2.1 \cite{murlabadia2026vjepa21} is our initialization baseline. In Tables~C1--C2, only the two DINOv2 \cite{oquab2024dinov2} models pretrained on LVD-142M are treated as gray scale references; the reproduced DINOv2 models trained on ImageNet-1K remain in the ranked comparison. In Table~C3, only the 7B-parameter DINOv3 \cite{simeoni2025dinov3} model is excluded as a scale reference.

\setcounter{table}{0}
\renewcommand{\thetable}{C\arabic{table}}
\renewcommand{\theHtable}{appendix.\arabic{table}}

\paragraph{Results.}
As shown in \Tabref{tab:segmentation}, Slot-JEPA consistently improves over its V-JEPA 2.1 initialization on all three segmentation datasets. The mIoU gains are 3.3\% on COCOStuff-27, 3.4\% on ADE20K, and 3.0\% on PASCAL VOC, with corresponding improvements in pixel accuracy; it also achieves the strongest results across all six metrics among the comparison models, excluding the gray reference rows. \Tabref{tab:depth} shows improvements on NYUv2 and on transfer from NYUv2 to SUNRGBD. On NYUv2, $\delta_1$ increases from 80.5\% to 83.1\% while AbsRel and RMSE decrease from 0.139 and 0.486 to 0.130 and 0.456. All three transfer metrics also improve, although performance on KITTI decreases slightly. Finally, \Tabref{tab:vos} reports an increase from 65.9\% to 66.6\% in DAVIS $\mathcal{J}\&\mathcal{F}$, with gains in both region similarity and contour accuracy. This is the strongest result among the comparison entries, supporting that slot reconstruction preserves temporally consistent object information.

\begin{table*}[!t]
\caption{Complete linear semantic-segmentation results (\%). ``Data'' and ``Ep.'' denote representation-pretraining data and epochs; IN1K, COCO, K400, and VM163M denote ImageNet-1K, COCO, Kinetics-400, and VisionMix-163M. $\dagger$ marks reproduced DINOv2.}
\label{tab:segmentation}
\centering
\small
\setlength{\tabcolsep}{2.5pt}
\begin{tabular*}{\textwidth}{@{\extracolsep{\fill}}lllcrrrrrr@{}}
\toprule
& & & & \multicolumn{2}{c}{COCOStuff-27} & \multicolumn{2}{c}{ADE20K} & \multicolumn{2}{c}{PASCAL VOC} \\
\cmidrule(lr){5-6}\cmidrule(lr){7-8}\cmidrule(l){9-10}
Method & Backbone & Data & Ep. & mIoU$\uparrow$ & Acc.$\uparrow$ & mIoU$\uparrow$ & Acc.$\uparrow$ & mIoU$\uparrow$ & Acc.$\uparrow$ \\
\midrule
\color{gray}DINOv2 & \color{gray}ViT-S/14 & \color{gray}LVD-142M & \color{gray}-- & \color{gray}57.4 & \color{gray}77.7 & \color{gray}43.5 & \color{gray}78.9 & \color{gray}79.4 & \color{gray}95.3 \\
\color{gray}DINOv2 & \color{gray}ViT-B/14 & \color{gray}LVD-142M & \color{gray}-- & \color{gray}59.1 & \color{gray}78.4 & \color{gray}46.3 & \color{gray}79.8 & \color{gray}81.3 & \color{gray}95.7 \\
\midrule
MoCo v2 & RN-50 & IN1K & 800 & 37.6 & 64.0 & 20.0 & 61.2 & 53.3 & 88.4 \\
SwAV & RN-50 & IN1K & 800 & 42.2 & 68.0 & 23.3 & 65.8 & 55.4 & 89.0 \\
DINO & RN-50 & IN1K & 800 & 40.7 & 66.6 & 22.4 & 64.5 & 53.1 & 88.4 \\
\midrule
MoCo v3 & ViT-S/16 & IN1K & 300 & 43.4 & 70.6 & 24.4 & 70.5 & 61.7 & 91.4 \\
DINO & ViT-S/16 & IN1K & 800 & 39.5 & 67.1 & 23.1 & 67.4 & 50.0 & 88.4 \\
iBOT & ViT-S/16 & IN1K & 800 & 47.7 & 72.7 & 28.5 & 72.3 & 66.7 & 92.4 \\
Mugs & ViT-S/16 & IN1K & 800 & 48.7 & 72.6 & 30.9 & 72.8 & 71.8 & 93.3 \\
Leopart & ViT-S/16 & IN1K+COCO & 800+50 & 48.8 & 73.2 & 29.7 & 72.1 & 68.2 & 92.6 \\
DINOv2\textsuperscript{$\dagger$} & ViT-S/14 & IN1K & 800 & 48.1 & 72.7 & 29.6 & 72.6 & 68.3 & 92.7 \\
\midrule
CAE & ViT-B/16 & IN1K & 800 & 45.5 & 71.0 & 27.0 & 71.9 & 64.9 & 92.2 \\
CAE & ViT-B/16 & IN1K & 1600 & 46.2 & 71.6 & 27.5 & 72.2 & 65.6 & 92.3 \\
MAE & ViT-B/16 & IN1K & 800 & 37.9 & 65.5 & 20.6 & 66.2 & 52.3 & 88.8 \\
MoCo v3 & ViT-B/16 & IN1K & 300 & 48.2 & 73.0 & 29.6 & 72.9 & 69.6 & 92.9 \\
DINO & ViT-B/16 & IN1K & 800 & 45.7 & 71.5 & 28.8 & 72.0 & 63.6 & 91.6 \\
DINO & ViT-B/8 & IN1K & 800 & 47.0 & 72.3 & 30.5 & 72.7 & 65.3 & 92.3 \\
iBOT & ViT-B/16 & IN1K & 400 & 52.2 & 75.3 & 34.3 & 75.2 & 73.3 & 93.9 \\
Mugs & ViT-B/16 & IN1K & 400 & 51.4 & 74.1 & 34.6 & 74.5 & 72.7 & 93.5 \\
DINOv2\textsuperscript{$\dagger$} & ViT-B/14 & IN1K & 400 & 53.0 & 75.6 & 35.5 & 75.8 & 74.9 & 93.9 \\
V-JEPA 2.1 & ViT-B/16 & VM163M & -- & 52.8 & 75.8 & 35.5 & 76.0 & 72.1 & 93.8 \\
\midrule
\textbf{Slot-JEPA} & ViT-B/16 & K400 & 10 & \textbf{56.1} & \textbf{77.8} & \textbf{38.9} & \textbf{77.8} & \textbf{75.1} & \textbf{94.5} \\
\bottomrule
\end{tabular*}
\end{table*}
\begin{table*}[!t]
\caption{Complete monocular-depth and NYUv2-to-SUNRGBD transfer results. $\dagger$ marks reproduced DINOv2.}
\label{tab:depth}
\centering
\small
\setlength{\tabcolsep}{1.5pt}
\begin{tabular*}{\textwidth}{@{\extracolsep{\fill}}llrrrrrrrrr@{}}
\toprule
& & \multicolumn{3}{c}{NYUv2} & \multicolumn{3}{c}{KITTI} & \multicolumn{3}{c}{NYU$\rightarrow$SUNRGBD} \\
\cmidrule(lr){3-5}\cmidrule(lr){6-8}\cmidrule(l){9-11}
Method & Backbone & $\delta_1\uparrow$ & AbsRel$\downarrow$ & RMSE$\downarrow$ & $\delta_1\uparrow$ & AbsRel$\downarrow$ & RMSE$\downarrow$ & $\delta_1\uparrow$ & AbsRel$\downarrow$ & RMSE$\downarrow$ \\
\midrule
\color{gray}DINOv2 (LVD-142M) & \color{gray}ViT-S/14 & \color{gray}84.8 & \color{gray}0.134 & \color{gray}0.437 & \color{gray}91.0 & \color{gray}0.096 & \color{gray}3.119 & \color{gray}77.4 & \color{gray}0.269 & \color{gray}0.473 \\
\color{gray}DINOv2 (LVD-142M) & \color{gray}ViT-B/14 & \color{gray}87.8 & \color{gray}0.120 & \color{gray}0.408 & \color{gray}92.0 & \color{gray}0.092 & \color{gray}2.931 & \color{gray}80.7 & \color{gray}0.255 & \color{gray}0.443 \\
\midrule
MoCo v2 & RN-50 & 62.9 & 0.230 & 0.719 & 82.4 & 0.130 & 4.319 & 55.5 & 0.366 & 0.749 \\
SwAV & RN-50 & 61.9 & 0.225 & 0.721 & 84.3 & 0.123 & 3.985 & 55.0 & 0.367 & 0.753 \\
DINO & RN-50 & 61.6 & 0.228 & 0.726 & 83.7 & 0.127 & 4.068 & 54.7 & 0.370 & 0.755 \\
\midrule
MoCo v3 & ViT-S/16 & 69.8 & 0.192 & 0.615 & 77.2 & 0.170 & 5.007 & 50.6 & 0.411 & 0.815 \\
DINO & ViT-S/16 & 67.7 & 0.202 & 0.651 & 83.9 & 0.132 & 4.061 & 59.6 & 0.343 & 0.678 \\
iBOT & ViT-S/16 & 72.1 & 0.183 & 0.584 & 85.3 & 0.125 & 3.907 & 63.5 & 0.330 & 0.634 \\
Mugs & ViT-S/16 & 72.3 & 0.179 & 0.586 & 84.1 & 0.132 & 3.963 & 64.4 & 0.306 & 0.613 \\
DINOv2\textsuperscript{$\dagger$} & ViT-S/14 & 72.7 & 0.184 & 0.583 & 86.6 & 0.112 & 3.757 & 63.2 & 0.332 & 0.631 \\
\midrule
MAE & ViT-B/16 & 65.3 & 0.224 & 0.673 & 84.3 & 0.123 & 4.087 & 59.1 & 0.350 & 0.688 \\
MoCo v3 & ViT-B/16 & 72.5 & 0.181 & 0.585 & 85.3 & 0.122 & 3.954 & 65.9 & 0.308 & 0.606 \\
DINO & ViT-B/16 & 71.7 & 0.182 & 0.593 & 86.5 & 0.117 & 3.771 & 62.8 & 0.336 & 0.644 \\
DINO & ViT-B/8 & 72.8 & 0.178 & 0.579 & 86.9 & 0.115 & 3.697 & 65.1 & 0.314 & 0.613 \\
iBOT & ViT-B/16 & 77.7 & 0.160 & 0.523 & 88.3 & 0.110 & 3.514 & 69.1 & 0.299 & 0.565 \\
Mugs & ViT-B/16 & 74.9 & 0.164 & 0.550 & 87.7 & 0.111 & 3.620 & 66.6 & 0.294 & 0.585 \\
DINOv2\textsuperscript{$\dagger$} & ViT-B/14 & 78.1 & 0.160 & 0.519 & 88.0 & 0.112 & 3.502 & 68.0 & 0.310 & 0.570 \\
V-JEPA 2.1 & ViT-B/16 & 80.5 & 0.139 & 0.486 & \textbf{89.2} & \textbf{0.102} & \textbf{3.481} & 70.0 & 0.292 & 0.557 \\
\midrule
\textbf{Slot-JEPA} & ViT-B/16 & \textbf{83.1} & \textbf{0.130} & \textbf{0.456} & 87.8 & 0.107 & 3.705 & \textbf{72.3} & \textbf{0.278} & \textbf{0.524} \\
\bottomrule
\end{tabular*}
\end{table*}
\clearpage
\begin{table}[H]
\caption{Complete DAVIS video object segmentation results (\%). MoCo v3, SwAV, and MAE report best-checkpoint $\mathcal{J}\&\mathcal{F}$ only.}
\label{tab:vos}
\centering
\small
\setlength{\tabcolsep}{5pt}
\begin{tabular}{llrrr}
\toprule
Method & Backbone & $\mathcal{J}\uparrow$ & $\mathcal{F}\uparrow$ & $\mathcal{J}\&\mathcal{F}\uparrow$ \\
\midrule
\color{gray}DINOv3 & \color{gray}ViT-7B/16 & \color{gray}-- & \color{gray}-- & \color{gray}71.1 \\
\midrule
MoCo v3 & ViT-S/16 & -- & -- & 61.7 \\
SwAV & RN-50 & -- & -- & 41.3 \\
DINO & ViT-B/16 & 61.5 & 64.9 & 63.2 \\
iBOT & ViT-B/16 & 63.2 & 65.5 & 64.3 \\
MAE & ViT-S/16 & -- & -- & 48.9 \\
CAE & ViT-L/16 & 62.8 & 67.5 & 65.1 \\
iBOT & ViT-L/16 & 62.1 & 65.5 & 63.8 \\
Mugs & ViT-L/16 & 60.3 & 63.0 & 62.1 \\
DINOv2 & ViT-L/14 & 62.2 & 66.3 & 64.2 \\
V-JEPA 2.1 & ViT-B/16 & 63.4 & 68.3 & 65.9 \\
\midrule
\textbf{Slot-JEPA} & ViT-B/16 & \textbf{63.5} & \textbf{69.7} & \textbf{66.6} \\
\bottomrule
\end{tabular}
\end{table}

\begin{figure}[H]
\centering
\includegraphics[width=\linewidth,height=0.64\textheight,keepaspectratio]{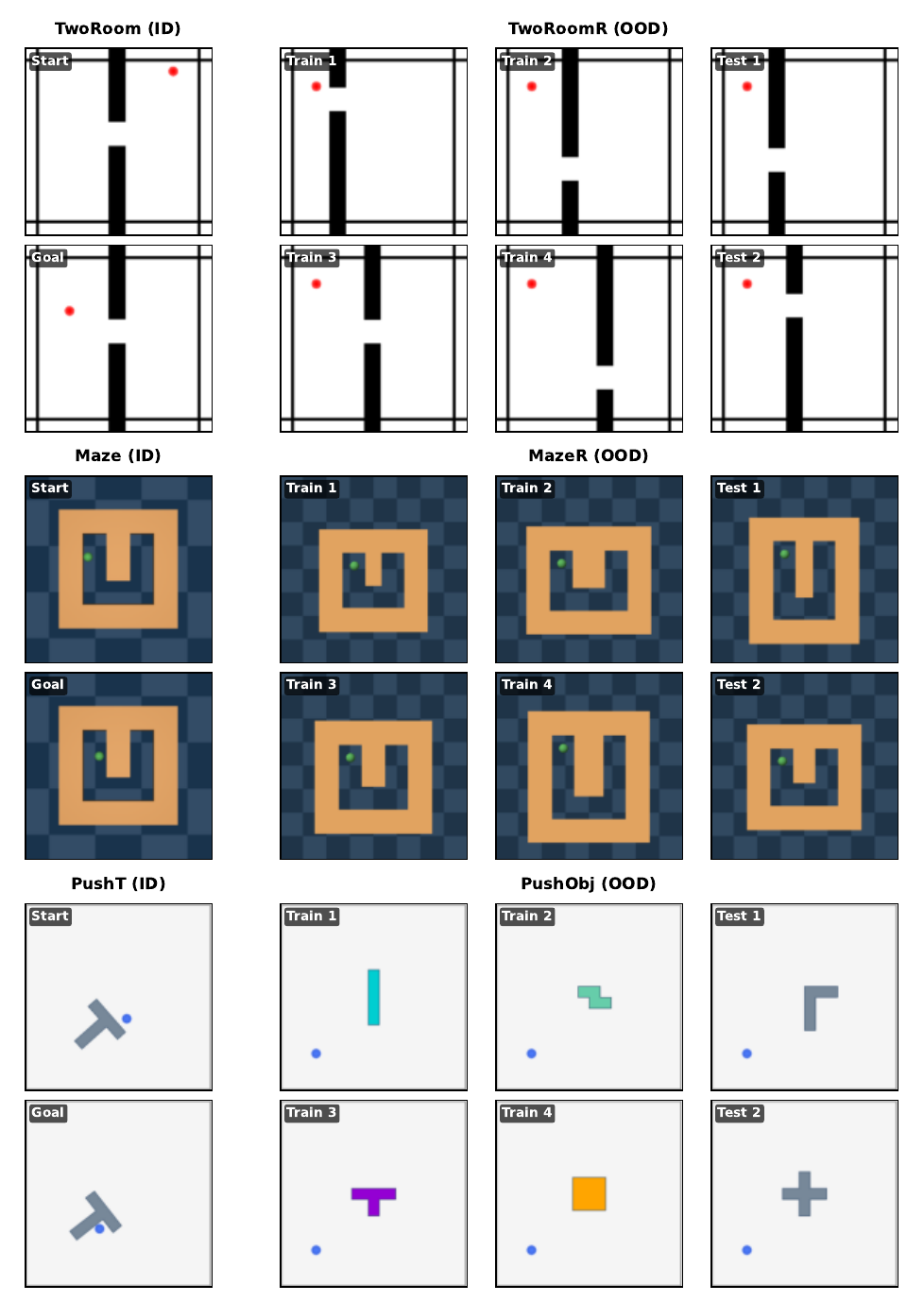}
\caption{The three planar ID/OOD environment pairs.}
\label{fig:planar-environment-examples}
\end{figure}

\subsection{Planar Planning}
\label{sec:appendix-planar}

\paragraph{Environment examples.}
\Figref{fig:planar-environment-examples} organizes the six environments into three ID/OOD pairs. Each pair forms a $2\times4$ grid of equally sized images: the first column contains the ID start and goal, the middle two columns contain four OOD training instances, and the final column contains two held-out OOD test instances. For TwoRoomR and MazeR, these are representative subsets of the training and test layouts; PushObj shows its complete shape split.

PushObj is trained with turquoise I, aquamarine Z, violet small-T, and orange square objects, and evaluated on the held-out L and plus shapes. Object poses vary within both splits; shape identity defines the ID/OOD distinction.

\paragraph{Competitors.}
The planar baselines use LeWorldModel (LeWM) \cite{maes2026leworldmodel}, DINO-WM \cite{zhou2025dinowm}, V-JEPA 2.1 \cite{murlabadia2026vjepa21}, DINO class-token representations \cite{caron2021dino}, ResNet \cite{he2016resnet}, and R3M \cite{nair2023r3m}. We report each method with and without the low-dimensional goal state when those results are available.

\paragraph{Slot and memory initialization.}
All planar cases use one fixed slot bank extracted from a small subset of training trajectories. In-distribution cases initialize the transition memory with randomly sampled training transitions. For an OOD case, we first execute random actions in the new simulated layout or with the new object shape and use the resulting transitions to initialize its memory. These observations are collected before the evaluated goal-directed rollout and decoded with the same fixed slots. Each transition subsequently observed during planning is appended to the bank.

\paragraph{Downstream fine-tuning.}
We freeze the visual encoder and slot decomposition module and optimize TMP with task-specific state and action encoders using AdamW. The learning rate is $5\times10^{-4}$ for both transferred and new predictor parameters, using the AdamW defaults $(\beta_1,\beta_2)=(0.9,0.999)$ and $\epsilon=10^{-8}$; no learning-rate schedule is applied. Training uses bfloat16, seed 0, and at most 1,000 iterations per epoch. Each training query predicts $\bm{M}_{t+1}$ from $(\bm{M}_t,r_t,a_t)$ using five observed context transitions from the same sampled training segment. During planning, five context transitions ($|\mathcal{C}_t|=5$) are sampled from the 50 nearest memory candidates. We set $(\lambda_{\mathrm{uot}},\lambda_{\mathrm{mask}},\lambda_{\mathrm{feat}})=(1,10,1)$ in $\mathcal{L}_{\mathrm{pred}}$, except for Push-Object, where we use $(0.1,10,1)$ together with a Sobel mask-boundary term of weight 10. UOT uses five Sinkhorn iterations, $\varepsilon=0.01$, and $\rho=0.2$.

\paragraph{Obstacle-aware UOT and MPC.}
For TwoRoom, TwoRoomR, Maze, and MazeR, the static and moving slot sets are estimated once from 128 deterministic training-transition pairs using the change score in Appendix~\ref{sec:appendix-uot-computation}; the resulting classification is then shared across test cases. We use the ten largest patch changes per transition, with $\tau_{\mathrm{static}}=0.05$ and $\tau_{\mathrm{moving}}=0.15$. Their UOT ground cost penalizes paths through the static-mask obstacle map. PushT and PushObj use the standard Euclidean ground cost. For each MPC step, the sampled memory contexts are fixed across CEM candidates. TMP rolls out every candidate action sequence, CEM ranks it with the UOT and mask-$\ell_1$ goal loss, and the agent executes the first action before observing the next state and replanning.

\begin{table}[!t]
\caption{Training settings for the planar environments.}
\label{tab:planar-training}
\centering
\small
\begin{tabular}{lrrrr}
\toprule
Training environment & Resolution & Epochs & Batch size & Weight decay \\
\midrule
TwoRoom / Maze & 384 & 100 & 128 & $10^{-5}$ \\
PushT & 512 & 100 & 128 & $10^{-5}$ \\
TwoRoomR / MazeR & 384 & 50 & 128 & $10^{-2}$ \\
PushObj & 512 & 500 & 64 & $10^{-4}$ \\
\bottomrule
\end{tabular}
\end{table}

\subsection{Robotic Arm Planning}
\label{sec:appendix-robot}

\paragraph{Environment and benchmark.}
We construct \textbf{REALM-JEPA} from the REALM simulator and task configurations \cite{sedlacek2025realm} to evaluate visual-goal planning in tabletop manipulation. The environment uses a mounted Franka Panda arm with a Robotiq gripper. At every step, it synchronously renders $1280\!\times\!720$ RGB observations from two fixed external cameras and one wrist-mounted camera. The robot state comprises a 6-D end-effector (EE) pose and the gripper opening, and the 7-D action specifies an absolute EE target pose and gripper command. REALM-JEPA contains 108 cases across four task families. \textit{Reach} includes 18 targets corresponding to the six Pick objects, six Rotate objects, and six objects used in Place. Each of the other families contains six task definitions under five scene configurations---the nominal scene, two shifted target-object poses, a block-distractor scene, and a shifted-distractor scene---giving 30 Pick, 30 Rotate, and 30 Place cases. Pick uses a banana, bowl, soda can, lemon, marker, or spoon; Rotate uses a banana, soda can, marker, mug, toy die, or water bottle; and Place uses the six object--receptacle pairs banana--box, soda can--bowl, green block--bowl, lemon--tray, marker--bowl, and orange--bowl. In total, the benchmark covers 11 manipulated-object types and three receptacle types, with $18+3\times(6\times5)=108$ cases.

\begin{figure}[!t]
\centering
\includegraphics[width=\linewidth]{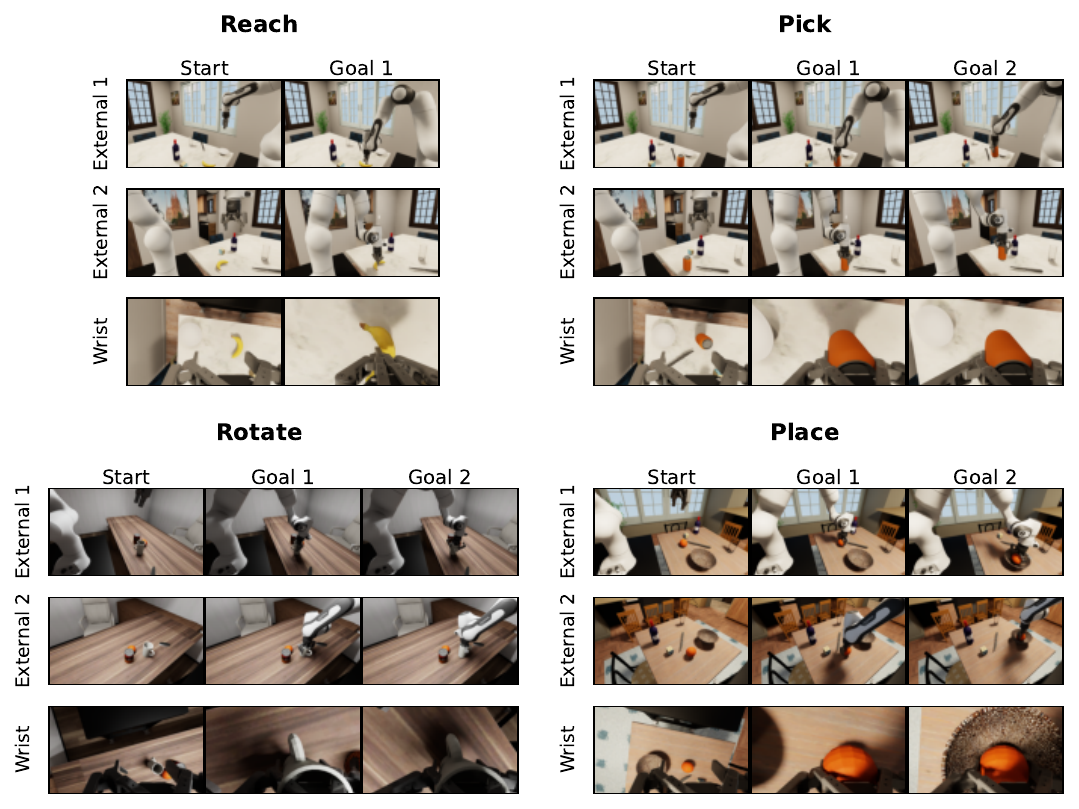}
\caption{Representative REALM-JEPA start and goal images for all four Robotic Arm Planning categories. Each task panel shows the same scene and goal stages from the two external cameras and the wrist camera. Goal 1 and Goal 2 denote successive visual goals when a task has multiple stages.}
\label{fig:realm-task-examples}
\end{figure}

Each benchmark case fixes the task configuration and initial simulator state, and provides synchronized visual goals from all three cameras. These goals are sampled from a scripted successful reference trajectory; the reference actions and intermediate states are used only to construct the benchmark and are unavailable to the planner. \Figref{fig:realm-task-examples} shows representative start and goal observations for every task family.
Reach moves the EE to the target object and uses one visual goal. Pick reaches and grasps the object before lifting it, Rotate grasps and changes its orientation, and Place grasps, transports, and releases it into the designated receptacle; these longer tasks use two successive visual goals. Every goal contains the same three synchronized views as the current observation.

\paragraph{Competitors.}
We compare with the standard V-JEPA 2.1 predictor \cite{murlabadia2026vjepa21} and V-JEPA 2.1 + TMP, which keeps patch representations and the original goal loss but uses our transition-memory architecture. This isolates the effect of the predictor design from the slot-based state interface. On place and rotate tasks, both patch-based baselines use the ground-truth target EE position to bias CEM proposals while the 3-D positional distance exceeds 30\,cm. Specifically, the proposal mean is initialized with 20\% of the target displacement and clipped per axis to 20\% of the action cap. At 30\,cm or closer, the bias is disabled and planning uses the visual goal loss alone. Slot-JEPA does not use this guidance.

\paragraph{Slot and memory initialization.}
Each REALM-JEPA case begins with a short random-action warmup. The initial frame, provided goal images, and warmup observations form the reference set for each view's slot extraction; the observed warmup transitions initialize its corresponding memory bank. Goal images contribute reference features but no memory transitions. Goal-directed planning starts only after this initialization, and each newly observed transition is appended to the appropriate per-view bank. After an extended rollout, slots may be refreshed from accumulated observed frames. All retained transitions are then decoded again so that their memory masks remain aligned with the refreshed slots. No future frame from the evaluated rollout is used for initialization or refresh.

\paragraph{Downstream fine-tuning.}
The robot predictor is trained on seven-frame DROID \cite{khazatsky2024droid} clips sampled at 1 fps and resized to $384\times384$. Training uses eight GPUs, four clips per GPU, bfloat16 precision, seed 239, 100 epochs, and 300 iterations per epoch. The visual encoder and slot decomposition module remain frozen, while TMP and the task-specific state and action encoders are optimized with AdamW using $(\beta_1,\beta_2)=(0.9,0.999)$, $\epsilon=10^{-8}$, and weight decay $10^{-2}$; bias and normalization parameters are excluded from weight decay. The learning rate increases linearly from $10^{-4}$ to $10^{-3}$ over the first 15 epochs, remains at $10^{-3}$ for 70 epochs, and decays linearly to zero over the final 15 epochs. We set $(\lambda_{\mathrm{uot}},\lambda_{\mathrm{mask}},\lambda_{\mathrm{feat}})=(3.0,3.0,0.1)$ in $\mathcal{L}_{\mathrm{pred}}$, with five Sinkhorn iterations and $\varepsilon=0.05$. Each training sample uses four observed frames to predict three future target frames and additionally applies two autoregressive prediction steps.

\paragraph{Multi-view prediction and planning.}
Let $v\in\mathcal{V}=\{\text{external-1},\text{external-2},\text{wrist}\}$ index the three views. They share the same predictor parameters $\theta$, but each view independently maintains its slot set $\bm{S}^{(v)}$, mask history, and transition memory $\mathcal{B}^{(v)}_t$. Context is sampled within each memory, and prediction is performed separately as
\begin{equation}
\hat{\bm{M}}^{(v)}_{t+1}=F_\theta\!\left(\bm{S}^{(v)},\bm{M}^{(v)}_t,r^{(v)}_t,a^{(v)}_t;\mathcal{C}^{(v)}_t\right).
\end{equation}
The same candidate robot action is transformed into each camera coordinate system using its camera pose before prediction; there is no cross-view attention or joint memory bank inside TMP. CEM then proceeds in two stages. First, both fixed views evaluate every pose candidate, and the mean of their goal losses updates the CEM mean and variance. Second, the wrist-view search is initialized with this updated distribution and continues optimizing the pose using the wrist-view goal loss. The best candidate from this refinement is executed; the gripper search is likewise evaluated from the wrist view. Thus the fixed cameras provide a stable global search direction, while the moving wrist camera refines the action from its local view.
Robot planning uses the standard UOT ground cost. Within each CEM stage, all candidates use the same sampled context transitions for the corresponding view.

\paragraph{Correlation distance.}
For the correlations in \Tabref{tab:robot-correlation}, EE--goal distance at step $t$ is
\begin{equation}
d^{\mathrm{EE}}_{t+1}=\left\|\bm{e}_{t+1}-\bm{e}_{g}\right\|_2,
\end{equation}
where $\bm{e}_{t+1}$ is the realized next 7-DoF state and $\bm{e}_{g}$ is the active visual-goal stage's target state; each contains the 6-D EE pose and gripper opening. This evaluation metric is distinct from the 3-D positional threshold used only for baseline guidance.

Predicted loss is the selected candidate's goal loss before executing action $a_t$, while observed loss is recomputed from the observation immediately after executing that action. Thus, predicted loss, observed loss, and $d^{\mathrm{EE}}_{t+1}$ refer to the same transition. To prevent task scale and trajectory length from dominating the aggregate, we standardize each variable within every task, view, and method, divide the distance range into 12 equal-width bins, and evenly subsample overrepresented bins. We compute each statistic per task and report the macro average across tasks.

\Tabref{tab:robot-correlation-full} reports all measured statistics for three comparisons: observed loss against physical distance, predicted loss against physical distance, and predicted loss against observed loss. Kendall's $\tau$ and Spearman's $\rho$ measure monotonic agreement. The standardized least-squares slope $\beta$ measures linear sensitivity, and $R^2$ measures linear goodness of fit. Slot-JEPA gives the strongest agreement for every comparison and camera view under all four statistics.

\begin{table}[!t]
\caption{Complete correlation analysis. Values are task-wise statistics macro-averaged after within-task standardization and distance-stratified sampling.}
\label{tab:robot-correlation-full}
\centering
\small
\setlength{\tabcolsep}{3pt}
\textbf{(a) Kendall's $\tau$}\\[-1pt]
\resizebox{0.9\linewidth}{!}{%
\begin{tabular}{@{}lccccccccc@{}}
\toprule
\raisebox{-7pt}[0pt][0pt]{Method} & \multicolumn{3}{c}{Obs. loss vs. EE--goal dist.}
& \multicolumn{3}{c}{Pred. loss vs. EE--goal dist.}
& \multicolumn{3}{c}{Pred. loss vs. Obs. loss} \\
\cmidrule(lr){2-4}\cmidrule(lr){5-7}\cmidrule(l){8-10}
& Ext. 1 & Ext. 2 & Wrist & Ext. 1 & Ext. 2 & Wrist & Ext. 1 & Ext. 2 & Wrist \\
\midrule
V-JEPA 2.1 & 0.299 & 0.326 & 0.397 & 0.386 & 0.345 & 0.248 & 0.391 & 0.481 & 0.419 \\
V-JEPA 2.1 + TMP & 0.167 & 0.075 & 0.538 & 0.277 & 0.266 & 0.356 & 0.322 & 0.304 & 0.599 \\
\textbf{Slot-JEPA} & \textbf{0.692} & \textbf{0.804} & \textbf{0.720} & \textbf{0.551} & \textbf{0.671} & \textbf{0.706} & \textbf{0.527} & \textbf{0.706} & \textbf{0.758} \\
\bottomrule
\end{tabular}
}

\vspace{4pt}
\textbf{(b) Spearman's $\rho$}\\[-1pt]
\resizebox{0.9\linewidth}{!}{%
\begin{tabular}{@{}lccccccccc@{}}
\toprule
\raisebox{-7pt}[0pt][0pt]{Method} & \multicolumn{3}{c}{Obs. loss vs. EE--goal dist.}
& \multicolumn{3}{c}{Pred. loss vs. EE--goal dist.}
& \multicolumn{3}{c}{Pred. loss vs. Obs. loss} \\
\cmidrule(lr){2-4}\cmidrule(lr){5-7}\cmidrule(l){8-10}
& Ext. 1 & Ext. 2 & Wrist & Ext. 1 & Ext. 2 & Wrist & Ext. 1 & Ext. 2 & Wrist \\
\midrule
V-JEPA 2.1 & 0.350 & 0.407 & 0.533 & 0.503 & 0.453 & 0.297 & 0.511 & 0.609 & 0.549 \\
V-JEPA 2.1 + TMP & 0.154 & 0.103 & 0.652 & 0.327 & 0.319 & 0.454 & 0.409 & 0.421 & 0.720 \\
\textbf{Slot-JEPA} & \textbf{0.792} & \textbf{0.891} & \textbf{0.807} & \textbf{0.677} & \textbf{0.788} & \textbf{0.814} & \textbf{0.668} & \textbf{0.825} & \textbf{0.865} \\
\bottomrule
\end{tabular}
}

\vspace{4pt}
\textbf{(c) Standardized slope $\beta$}\\[-1pt]
\resizebox{0.9\linewidth}{!}{%
\begin{tabular}{@{}lccccccccc@{}}
\toprule
\raisebox{-7pt}[0pt][0pt]{Method} & \multicolumn{3}{c}{Obs. loss vs. EE--goal dist.}
& \multicolumn{3}{c}{Pred. loss vs. EE--goal dist.}
& \multicolumn{3}{c}{Pred. loss vs. Obs. loss} \\
\cmidrule(lr){2-4}\cmidrule(lr){5-7}\cmidrule(l){8-10}
& Ext. 1 & Ext. 2 & Wrist & Ext. 1 & Ext. 2 & Wrist & Ext. 1 & Ext. 2 & Wrist \\
\midrule
V-JEPA 2.1 & 0.423 & 0.391 & 0.594 & 0.549 & 0.519 & 0.391 & 0.601 & 0.610 & 0.504 \\
V-JEPA 2.1 + TMP & 0.144 & 0.106 & 0.699 & 0.388 & 0.360 & 0.544 & 0.411 & 0.475 & 0.745 \\
\textbf{Slot-JEPA} & \textbf{0.865} & \textbf{0.940} & \textbf{0.874} & \textbf{0.706} & \textbf{0.848} & \textbf{0.882} & \textbf{0.730} & \textbf{0.887} & \textbf{0.922} \\
\bottomrule
\end{tabular}
}

\vspace{4pt}
\textbf{(d) Coefficient of determination $R^2$}\\[-1pt]
\resizebox{0.9\linewidth}{!}{%
\begin{tabular}{@{}lccccccccc@{}}
\toprule
\raisebox{-7pt}[0pt][0pt]{Method} & \multicolumn{3}{c}{Obs. loss vs. EE--goal dist.}
& \multicolumn{3}{c}{Pred. loss vs. EE--goal dist.}
& \multicolumn{3}{c}{Pred. loss vs. Obs. loss} \\
\cmidrule(lr){2-4}\cmidrule(lr){5-7}\cmidrule(l){8-10}
& Ext. 1 & Ext. 2 & Wrist & Ext. 1 & Ext. 2 & Wrist & Ext. 1 & Ext. 2 & Wrist \\
\midrule
V-JEPA 2.1 & 0.402 & 0.362 & 0.391 & 0.394 & 0.352 & 0.237 & 0.424 & 0.477 & 0.312 \\
V-JEPA 2.1 + TMP & 0.313 & 0.262 & 0.530 & 0.290 & 0.311 & 0.404 & 0.312 & 0.490 & 0.602 \\
\textbf{Slot-JEPA} & \textbf{0.792} & \textbf{0.890} & \textbf{0.793} & \textbf{0.531} & \textbf{0.740} & \textbf{0.793} & \textbf{0.561} & \textbf{0.797} & \textbf{0.859} \\
\bottomrule
\end{tabular}
}
\end{table}

\paragraph{Action execution.}
The 7-D planning command consists of a 6-D end-effector target pose and a scalar gripper-opening command. In absolute-pose mode, the arm controller computes Cartesian pose error, maps it through inverse kinematics to desired joint positions, and applies joint-space PD feedback with Jacobian-weighted Cartesian gains. A separate continuous gripper controller applies the opening command. We allow at most 100 MPC steps per planning attempt.

\subsection{Additional Visualizations}
\label{sec:appendix-visualizations}

\begin{figure}[t]
\centering
\includegraphics[width=\linewidth]{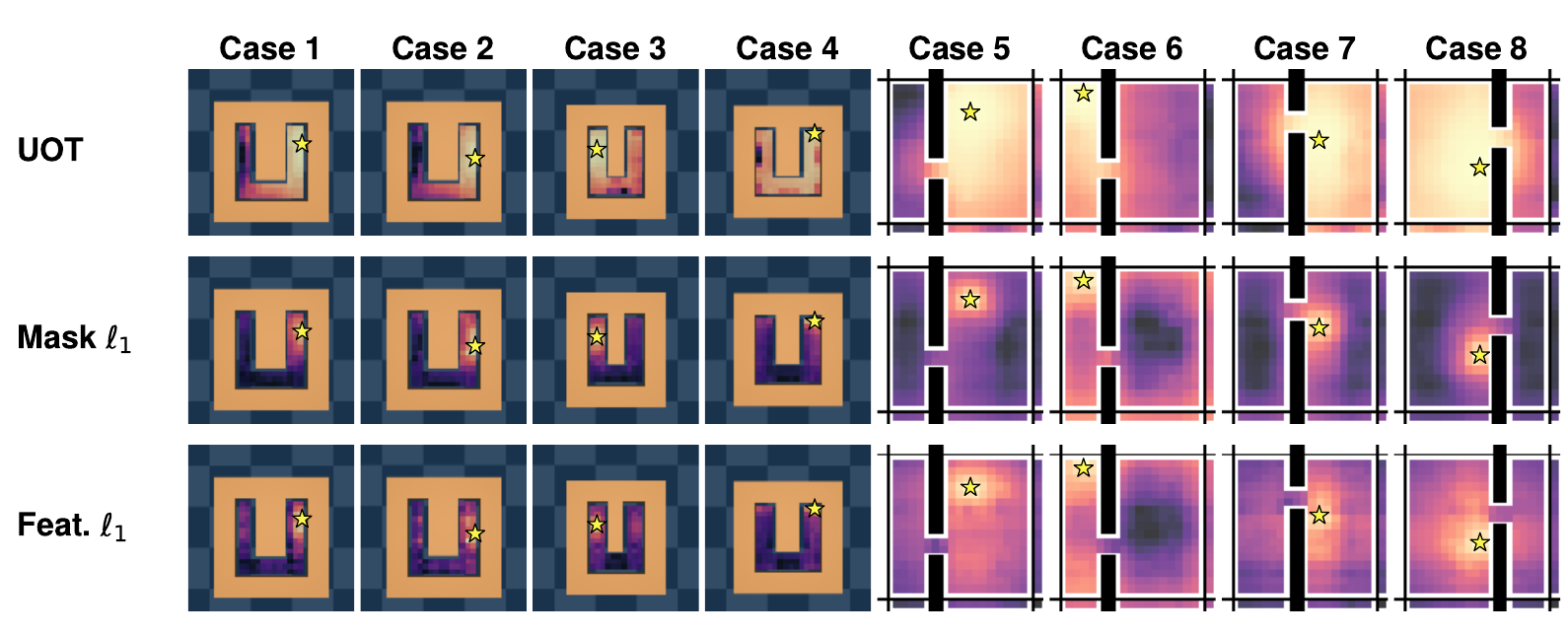}
\caption{Additional goal-loss landscapes in MazeR and TwoRoomR. Each sampled state is rendered independently; the panels visualize energy geometry rather than executed rollouts. Stars mark goals, and lighter colors indicate lower loss.}
\label{fig:appendix-loss-landscapes}
\end{figure}

\paragraph{UOT objective.}
\Figref{fig:appendix-loss-landscapes} shows how the goal loss varies over sampled states. Low-loss regions follow feasible progress toward the goal while the obstacle-aware ground cost assigns less favorable energy to visually nearby but obstructed states, yielding a planning signal that better reflects scene geometry.

\begin{figure}[t]
\centering
\includegraphics[width=\linewidth]{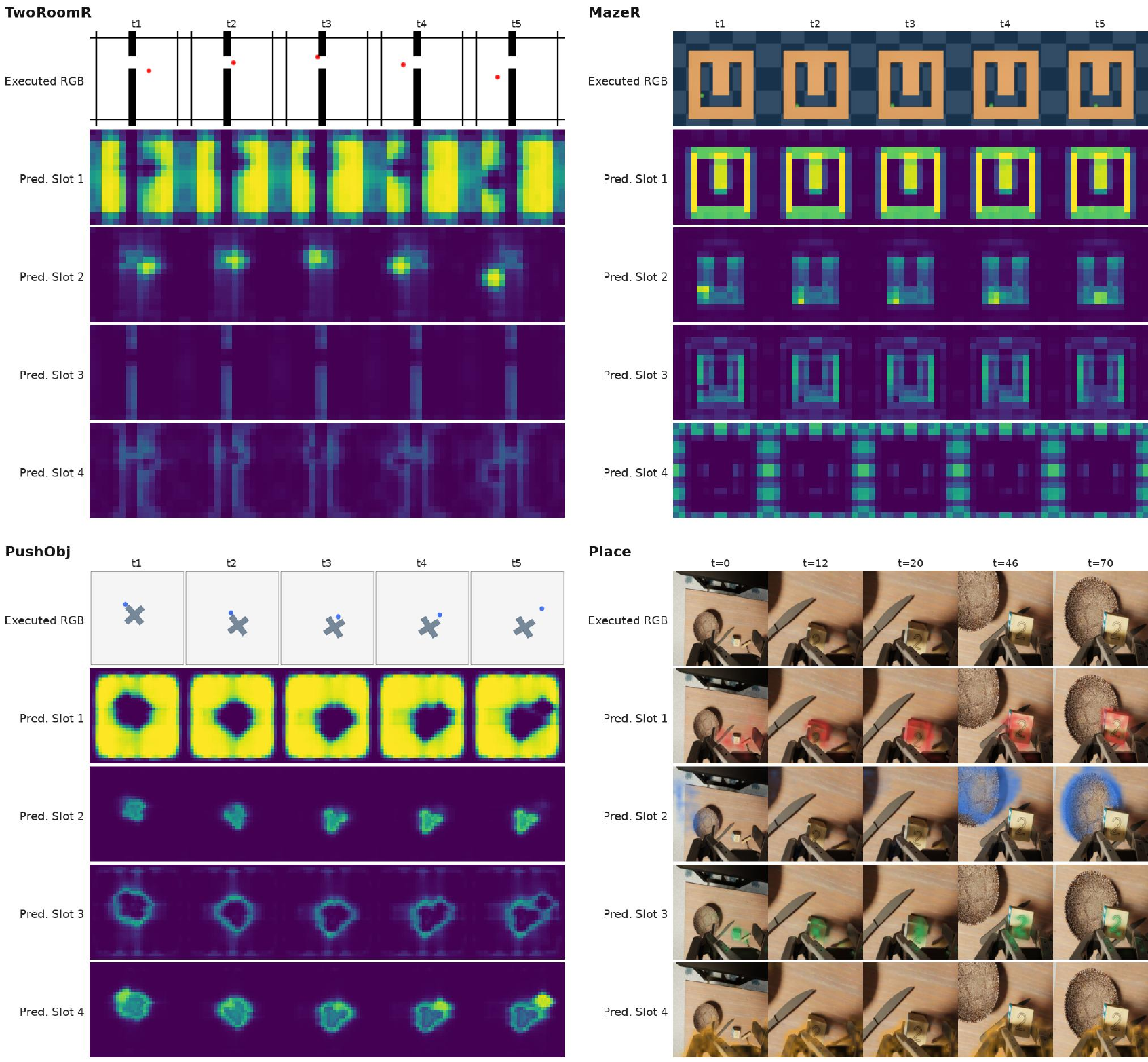}
\caption{Predicted slot masks along successful TwoRoomR, MazeR, PushObj, and Place planning rollouts (top left, top right, bottom left, and bottom right, respectively). Each column pairs the RGB observation obtained after action execution with masks predicted for the same time step. The planar panels show mask heatmaps, whereas the Place panel overlays sharpened predictions on the executed wrist-view RGB.}
\label{fig:appendix-planar-slot-rollouts}
\end{figure}

\paragraph{Predicted masks.}
\Figref{fig:appendix-planar-slot-rollouts} compares action-conditioned slot-mask predictions with the RGB observations obtained after execution. The predicted masks track the task-relevant regions across environments and time. Because the future is represented as separate slot masks, the prediction is also more interpretable: one can inspect which object or region is expected to move and where, rather than only comparing an opaque dense latent.

\begin{figure}[t]
\centering
\includegraphics[width=\linewidth,height=0.88\textheight,keepaspectratio]{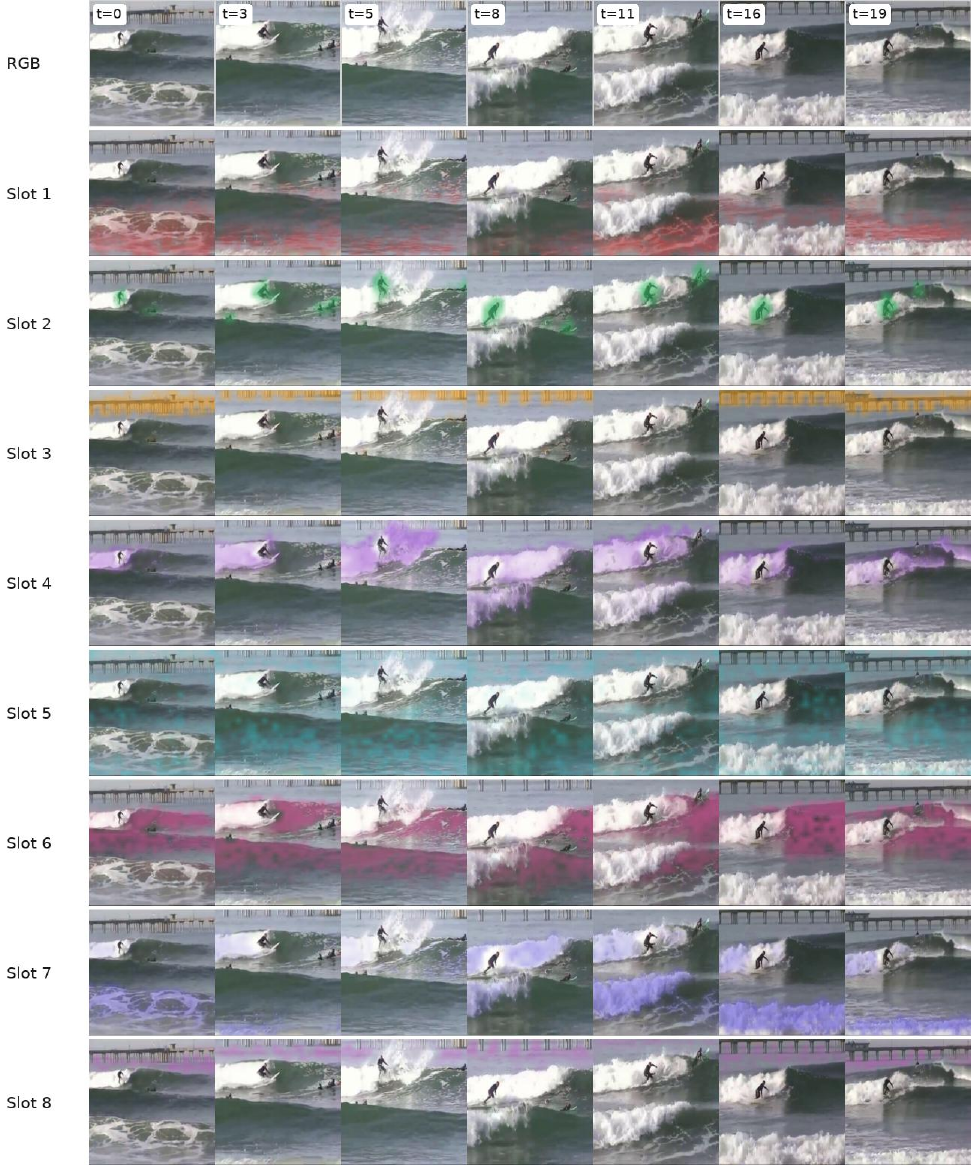}
\caption{Slot decomposition on Kinetics-400 pretraining example. The first row contains the RGB frames, and each subsequent row overlays one selected soft slot mask.}
\label{fig:appendix-slot-pretraining-2}
\end{figure}

\begin{figure}[t]
\centering
\includegraphics[width=\linewidth,height=0.88\textheight,keepaspectratio]{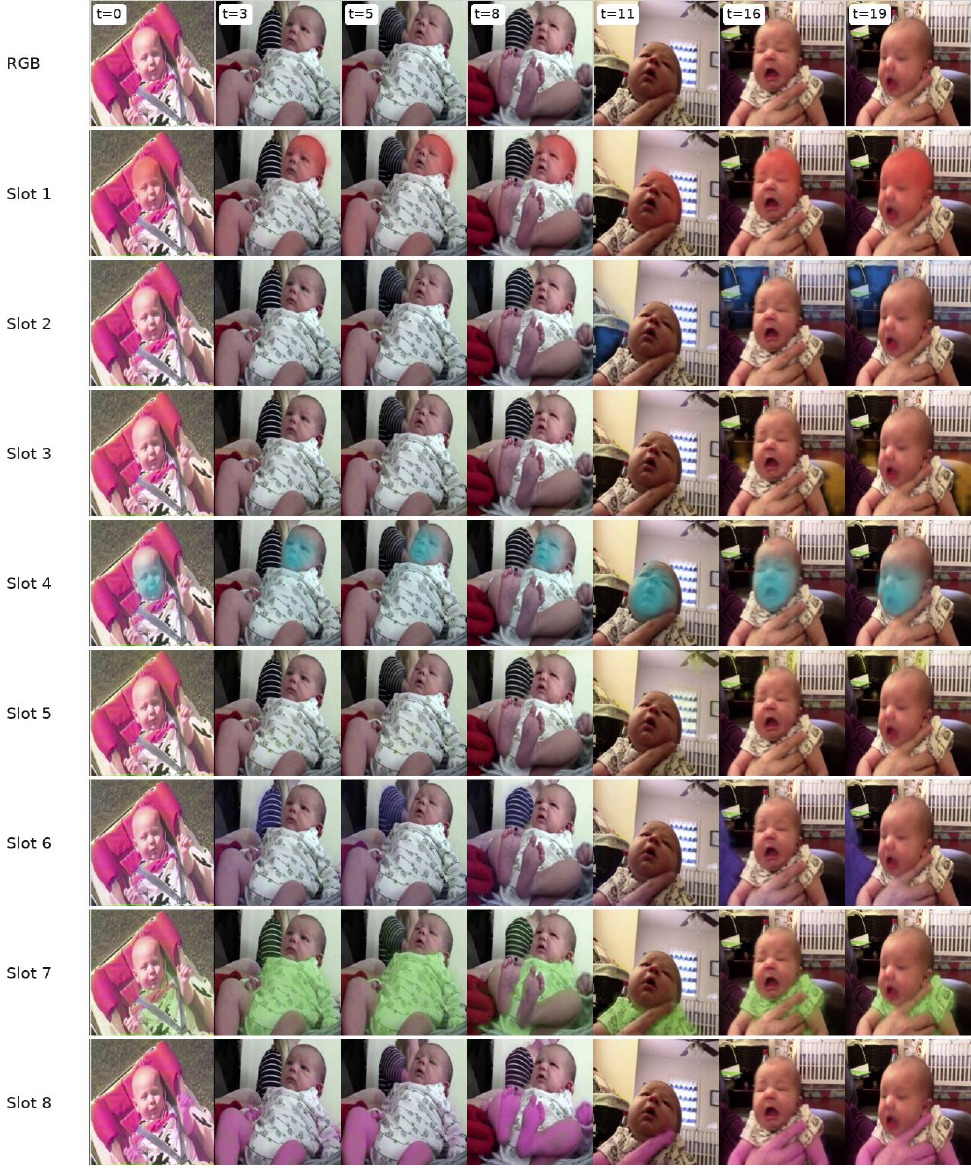}
\caption{Slot decomposition on Kinetics-400 pretraining example. The first row contains the RGB frames, and each subsequent row overlays one selected soft slot mask.}
\label{fig:appendix-slot-pretraining-6}
\end{figure}

\begin{figure}[t]
\centering
\includegraphics[width=\linewidth,height=0.88\textheight,keepaspectratio]{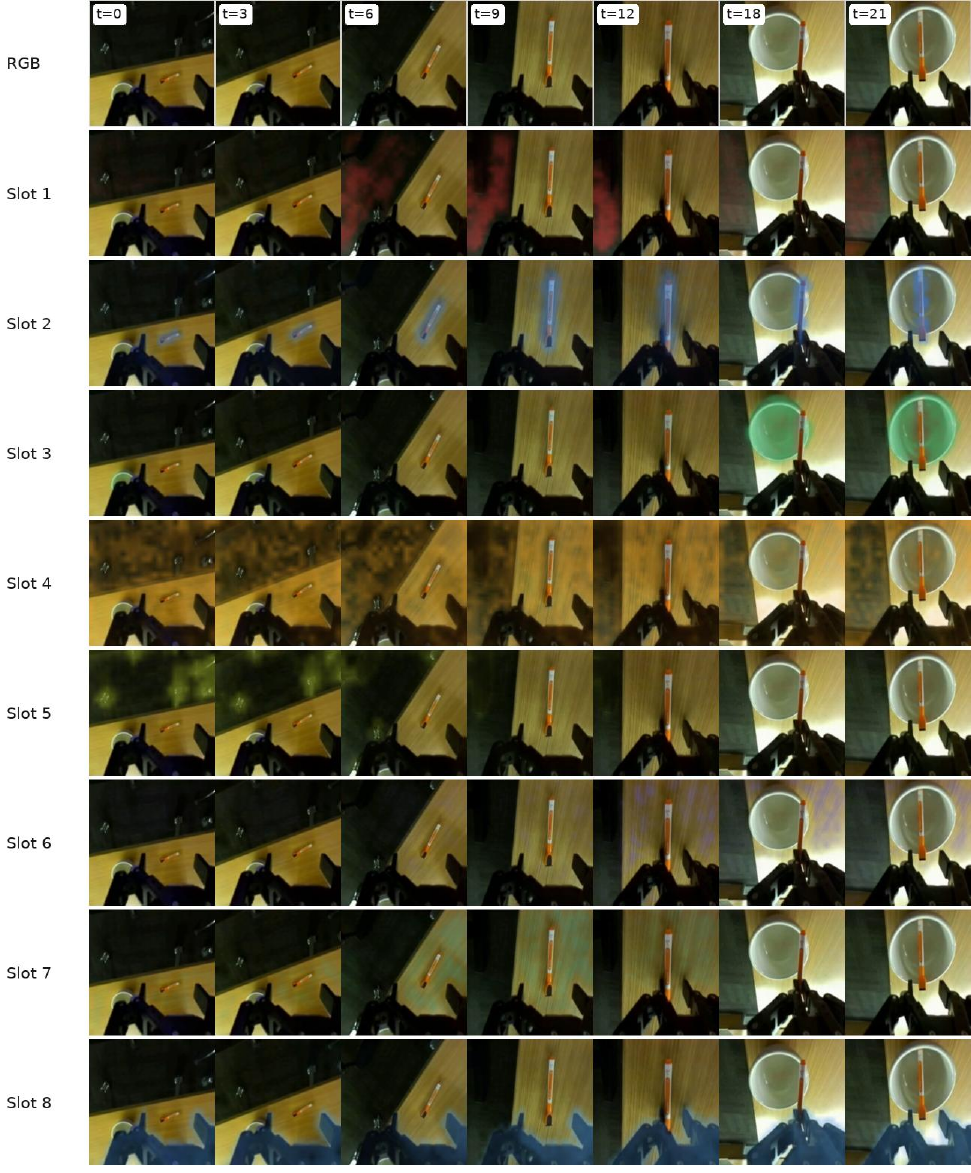}
\caption{Slot decomposition on DROID downstream-training example. The first row contains the RGB frames, and each subsequent row overlays one selected soft slot mask.}
\label{fig:appendix-slot-downstream-5}
\end{figure}

\begin{figure}[t]
\centering
\includegraphics[width=\linewidth,height=0.88\textheight,keepaspectratio]{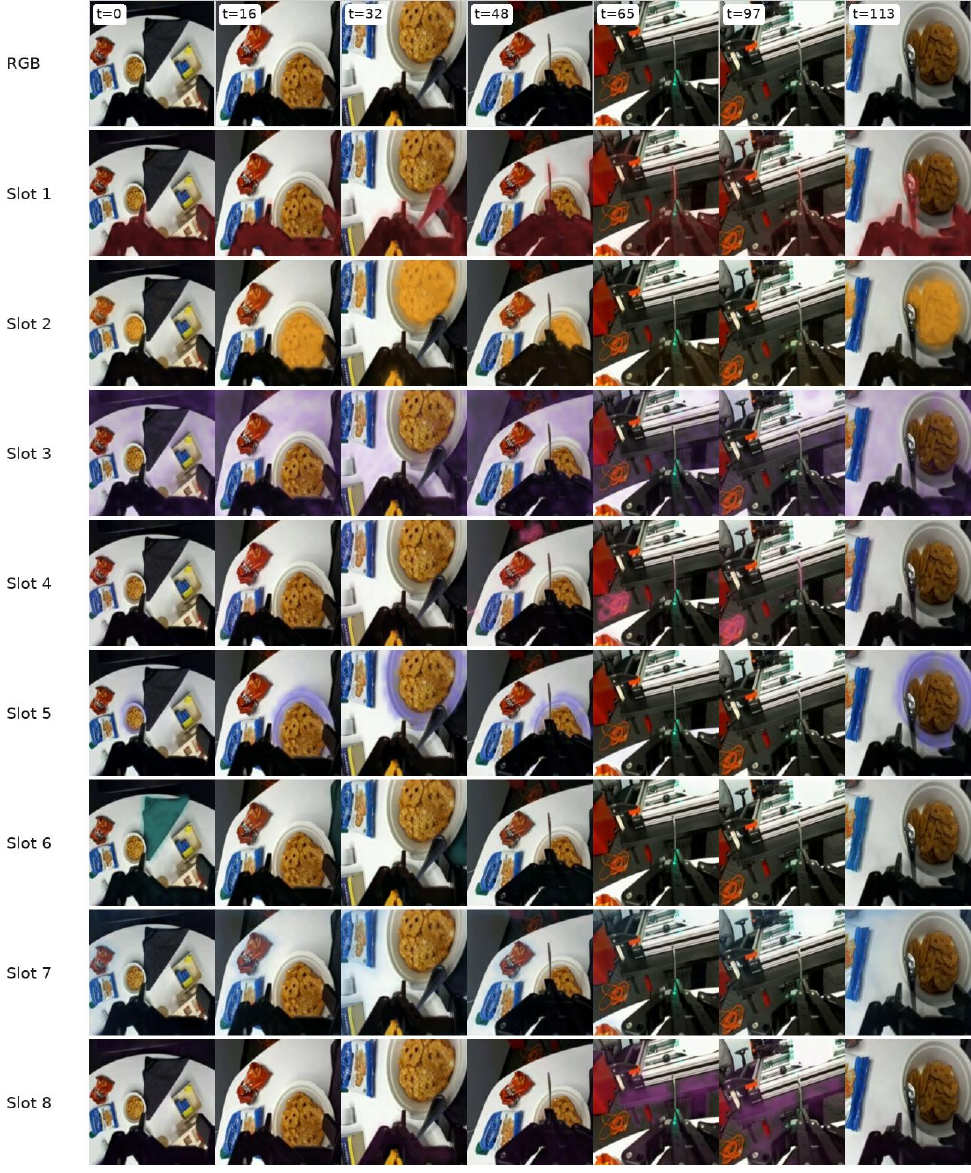}
\caption{Slot decomposition on DROID downstream-training example. The first row contains the RGB frames, and each subsequent row overlays one selected soft slot mask.}
\label{fig:appendix-slot-downstream-6}
\end{figure}

\paragraph{Slot decomposition.}
Figures~\ref{fig:appendix-slot-pretraining-2}--\ref{fig:appendix-slot-downstream-6} visualize individual slots as colored masks over the input RGB frames. Across Kinetics-400 pretraining videos and DROID downstream-training clips, different slots emphasize distinct objects or spatial regions and remain recognizable over time. This decomposition exposes which parts of a scene contribute to the latent state instead of collapsing them into one dense feature map.

Overall, these visualizations indicate that slot decomposition provides a spatially structured and inspectable state, while UOT and TMP turn that state into goal-directed costs and coherent future-mask predictions.

\flushbottom


\end{document}